\documentclass{article}

\usepackage[final, eandd]{neurips_2026}

\makeatletter
\renewcommand{\@notice}{}
\makeatother

\usepackage[utf8]{inputenc}
\usepackage[T1]{fontenc}
\usepackage{hyperref}
\usepackage{url}
\usepackage{booktabs}
\usepackage{amsfonts}
\usepackage{nicefrac}
\usepackage{microtype}
\usepackage[table]{xcolor}
\usepackage{graphicx}
\usepackage{subcaption}
\usepackage{amsmath}
\usepackage{amsthm}
\usepackage{mathtools}
\usepackage{xspace}
\usepackage{multirow}
\usepackage{enumitem}
\usepackage{algorithm}
\usepackage{algorithmic}
\usepackage{cleveref}
\usepackage{threeparttable}
\usepackage{wasysym}
\usepackage{tabularx}
\usepackage{bm}

\newcommand{\dataset}{\textsc{HiGraph}\xspace}

\newcolumntype{C}{>{\centering\arraybackslash}X}

\title{\dataset: A Large-Scale Hierarchical Graph Dataset for Malware Analysis}

\author{%
  \normalfont
  Han Chen\textsuperscript{1} \quad Hanchen Wang\textsuperscript{1} \quad Hongmei Chen\textsuperscript{2} \\
  Ying Zhang\textsuperscript{1} \quad Lu Qin\textsuperscript{1} \quad Wenjie Zhang\textsuperscript{3} \\[0.6ex]
  \textsuperscript{1}University of Technology Sydney, Sydney, Australia \\
  \textsuperscript{2}Yunnan University, Kunming, China \quad
  \textsuperscript{3}University of New South Wales, Sydney, Australia \\[0.4ex]
  \texttt{han.chen-7@student.uts.edu.au} \quad
  \texttt{\{hanchen.wang, ying.zhang, lu.qin\}@uts.edu.au} \\
  \texttt{hmchen@ynu.edu.cn} \quad \texttt{zhangw@cse.unsw.edu.au}
}

\begin{document}

\maketitle

\begin{abstract}
The advancement of graph-based malware analysis is critically limited by the absence of large-scale datasets that capture the inherent hierarchical structure of software. Existing methods often oversimplify programs into single-level graphs, failing to model the crucial semantic relationship between high-level functional interactions and low-level instruction logic. To bridge this gap, we introduce \dataset, the largest public hierarchical graph dataset for malware analysis, comprising over \textbf{200M} Control Flow Graphs (CFGs) nested within \textbf{499K} Function Call Graphs (FCGs). This two-level representation preserves structural semantics essential for building robust detectors resilient to code obfuscation and malware evolution. We demonstrate \dataset's utility through a large-scale analysis that reveals distinct structural properties of benign and malicious software, establishing it as a foundational benchmark for the community. The dataset and an interactive explorer are publicly available at \url{https://higraph.org}.
\end{abstract}

\section{Introduction}
Graph neural networks (GNNs) \cite{kipf2017semi,velivckovic2018graph,xu2018powerful,hamilton2017inductive} offer a promising frontier for malware analysis, as they can capture complex structural patterns resilient to common obfuscation techniques. However, the development of robust GNN-based detectors is severely hampered by the lack of large-scale, high-quality datasets. Existing malware corpora often suffer from critical limitations: they may not represent programs as graphs, or they possess temporal biases (e.g., training on "future" samples) that lead to unrealistic performance evaluations \cite{arp2022dos,pendlebury2019tesseract}. More fundamentally, even public graph-based datasets like Malnet~\cite{freitas2020large} typically represent programs as single-level, ``flat'' graphs. This oversimplification fails to capture the rich, hierarchical nature of software, where low-level control flow within functions and high-level interactions between them are both crucial for understanding program behavior.

Our dataset construction is driven by a critical insight: despite continuous surface-level evolution, the \textbf{core malicious behaviors} within a malware family remain remarkably stable. Ransomware like CryptoLocker, for instance, has changed its encryption details from XOR to hybrid AES+RSA, yet its workflow (\textit{file discovery} $\to$ \textit{encryption} $\to$ \textit{notification}) remains invariant. A hierarchical approach that combines Control Flow Graphs (CFGs) for intra-procedural logic and Function Call Graphs (FCGs) for inter-procedural interactions captures such structural invariants, enabling detection that transcends superficial code modifications~\cite{bilotSurveyMalwareDetection2023,he2022msdroid,zhangSemanticpreservingReinforcementLearning2022,yangCADEDetectingExplaining2021,zhang2020enhancing,zhangSemanticsAwareAndroidMalware2014,loGraphNeuralNetworkbased2022}.

To address these challenges and enable more sophisticated analysis of program hierarchies, we introduce \dataset, the largest publicly available hierarchical graph dataset for malware analysis. \dataset encompasses \textbf{499,981} applications, represented as \textbf{499,981} global FCGs that contain a total of \textbf{201,792,085} local CFGs. This two-level structure (illustrated in Figure~\ref{fig:higraph_overview}b alongside the construction pipeline) explicitly captures both inter-procedural dependencies and intra-procedural logic. By providing granular hierarchical data with spatio-temporal consistency, \dataset provides a robust foundation for building and evaluating the next generation of malware detectors. To facilitate exploration, we also provide an interactive website for the dataset at \textbf{\url{https://higraph.org}}. Empirically, our hierarchical Hi-GNN attains the strongest static IID PR-AUC ($0.785$) and Macro F1 ($0.761$), and sustains AUT(F1) of $0.755$ over 2012$\rightarrow$2013 and $0.715$ over 2012$\rightarrow$2016, surpassing flat-graph GNNs and the modern graph transformer GraphGPS~\cite{rampavsek2022recipe} on all metrics except short-term AUT(PR-AUC) under matched feature and hidden dimensions, with the discriminative signal localizing to CFG-level cyclomatic complexity (Cohen's $d{=}0.48$)---a structural property invisible to FCG-only architectures.

In summary, the contributions of this paper are as follows:
\begin{itemize}[topsep=0mm, itemsep=1mm, parsep=1mm, leftmargin=*]
  \item We release \dataset, the largest hierarchical graph dataset for Android malware analysis to date, comprising 499{,}981 applications with 499{,}981 FCGs and 201{,}792{,}085 nested CFGs, labeled at the VirusTotal $\geq$15-engine threshold and annotated with AVClass2~\cite{sebastian2020avclass2} families across 2012--2022.

  \item We define a longitudinal benchmark with three evaluation regimes (IID, short-term, and long-term) and a cross-anchor concept-drift protocol (Train 2012 and Train 2016) covering 433{,}488 test apps over 10 years.

  \item We benchmark classical GNNs and a modern graph transformer on \dataset under matched feature and hidden dimensions, showing that hierarchical inductive bias yields drift robustness that a stronger flat-graph architecture does not match in our setup.

  \item We open-source \dataset, the preprocessing pipeline, and the evaluation harness to standardize longitudinal evaluation of hierarchical malware analysis methods and foster reproducible research in AI for cybersecurity.
\end{itemize}

\section{Related Work}

\paragraph{Malware Datasets.}
Research in malware analysis is supported by numerous datasets. Large scale repositories like AndroZoo~\cite{allix2016androzoo} for Android and VirusShare~\cite{virusshare2025} for Windows provide vast collections of raw malware samples. Other datasets offer more structured information, such as API call graphs from APIGraph~\cite{zhang2020enhancing} or network traffic from CICMalDroid~\cite{mahdavifar2020dynamic}. However, among datasets designed for graph based analysis, MalNet~\cite{freitas2020large} is the most prominent. While extensive, MalNet represents programs as flat graphs, overlooking the inherent hierarchical structure of software. This structural simplification limits the potential for more nuanced analysis, creating a need for datasets that capture the multilevel organization of program code.

\paragraph{Graph Representation Learning.}
Graph representation learning has rapidly evolved from early node embedding methods~\cite{grover2016node2vec,perozzi2014deepwalk} to a variety of powerful Graph Neural Networks (GNNs), such as Graph Convolutional Networks (GCNs)~\cite{kipf2017semi}, Graph Attention Networks (GATs)~\cite{velivckovic2018graph}, and transformer based architectures~\cite{rampavsek2022recipe,rong2020self}. These models have demonstrated strong performance in malware classification~\cite{he2022msdroid,chen2023guided}. Nevertheless, their effectiveness is primarily on conventional, non-hierarchical graphs, making them unable to directly leverage the nested relationships present in complex systems like software.

\paragraph{Hierarchical Graph Learning.}
To model systems with nested structures, the concept of hierarchical graphs, or Graphs of Graphs (GoG), has emerged~\cite{li2019semi,chen2023denoising,wang2020gognn}. This paradigm has been applied in fields such as computational biology~\cite{wang2021multi,gao2023hierarchical} and for detecting malware on Windows~\cite{ling2022malgraph}. However, its application to Android malware has been unexplored, largely due to the absence of suitable datasets with explicit hierarchical information. To bridge this gap, we introduce \dataset, a new, large scale dataset designed specifically for hierarchical graph learning on Android applications. As detailed in Table~\ref{tab:dataset_comparison}, \dataset provides the first opportunity for the community to explore hierarchical representation learning for malware analysis at scale.

\begin{table}[t]
  \centering
  \begin{threeparttable}
    \small
    \setlength{\tabcolsep}{6pt}
    \begin{tabular}{@{}lcccccc@{}}
      \toprule
      & \multicolumn{2}{c}{\textbf{Dataset Properties}} & \multicolumn{2}{c}{\textbf{Graph Features}} & \multicolumn{2}{c}{\textbf{Quality Assurance}} \\
      \cmidrule(lr){2-3} \cmidrule(lr){4-5} \cmidrule(lr){6-7}
      \textbf{Dataset} & \textbf{Year} & \textbf{Size} & \textbf{Format} & \textbf{Scale} & \textbf{Temp.} & \textbf{Spat.} \\
      \midrule
      AndroZoo~\cite{allix2016androzoo}     & 2016 & 11M+              & Raw APKs     & \CIRCLE     & \CIRCLE     & \Circle \\
      Drebin~\cite{arp2014drebin}            & 2014 & 5.5K              & Features     & \Circle     & \LEFTcircle & \Circle \\
      AMD~\cite{wei2017deep}                 & 2017 & NA\tnote{1}       & Features     & \Circle     & \Circle     & \Circle \\
      APIGraph~\cite{zhang2020enhancing}     & 2020 & 322K              & Features     & \LEFTcircle & \CIRCLE     & \CIRCLE \\
      Malnet~\cite{freitas2020large}         & 2021 & 1.2M              & Single-level & \CIRCLE     & \Circle     & \Circle \\
      \midrule
      \dataset{} (Ours)                      & 2025 & 499K (200M+CFGs)  & Hierarchical & \CIRCLE     & \CIRCLE     & \CIRCLE \\
      \bottomrule
    \end{tabular}
    \begin{tablenotes}
      \scriptsize
      \item[a] \CIRCLE=high/full, \Circle=low/none, \LEFTcircle=medium/partial.
      \item[b] Scale indicates data scale size. Temp.=Temporal bias robustness, Spat.=Spatial bias robustness. Format refers to the data representation used by each dataset.
      \item[1] The dataset is no longer publicly available according to the authors.
    \end{tablenotes}
    \caption{Comparison with existing malware datasets in terms of dataset properties, graph features, and quality assurance.}
    \label{tab:dataset_comparison}
  \end{threeparttable}
\end{table}

\section{Dataset Construction}
\label{sec:construction}

\begin{figure}[t]
  \centering
  \begin{minipage}[b]{0.50\textwidth}
    \centering
    \includegraphics[width=\linewidth]{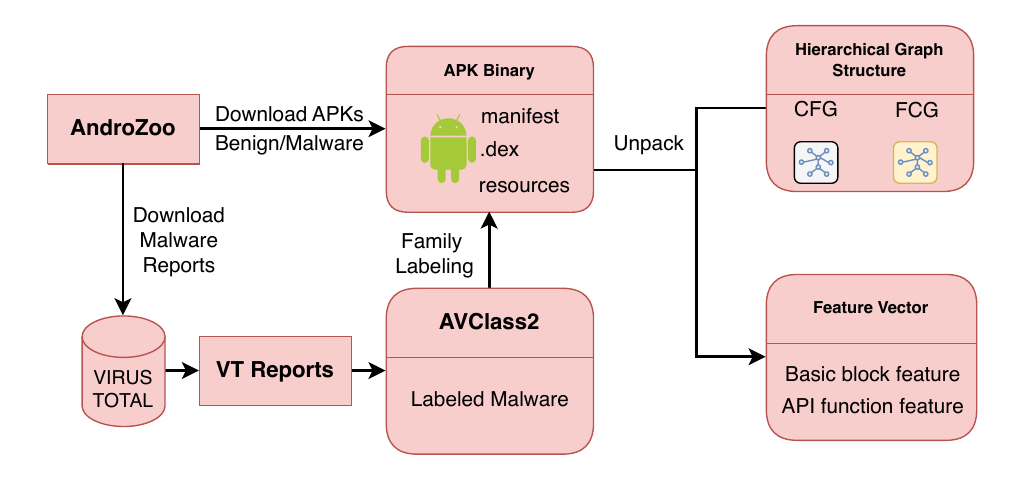}
    \subcaption{Construction pipeline. We download APKs from AndroZoo, label them with VirusTotal reports, assign families via AVClass2, and extract hierarchical CFG/FCG graphs with feature vectors.}
    \label{fig:procedure}
  \end{minipage}\hfill
  \begin{minipage}[b]{0.48\textwidth}
    \centering
    \includegraphics[width=\linewidth]{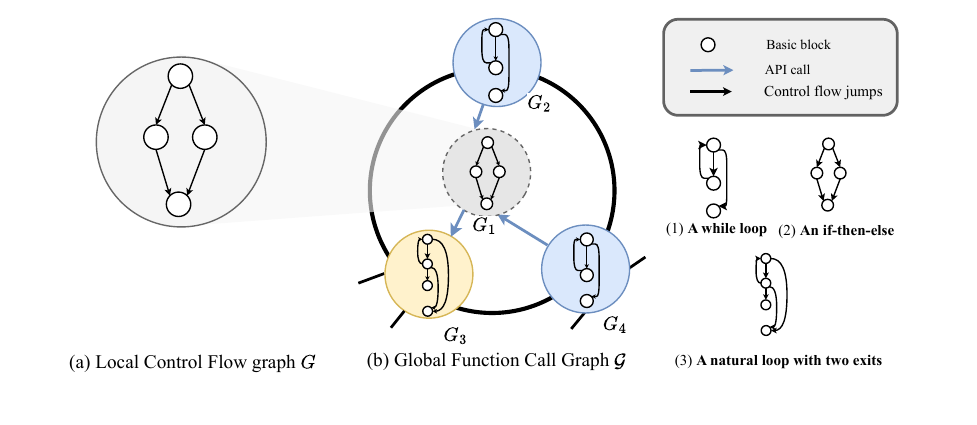}
    \subcaption{Resulting two-level structure. (left) A local CFG $G$ captures intra-procedural instruction flow within a function. (right) A global FCG $\mathcal{G}$ links functions ($G_1$--$G_4$) via API calls, exposing inter-procedural behavior.}
    \label{fig:higraph}
  \end{minipage}
  \caption{Overview of \dataset: (a) end-to-end construction pipeline; (b) two-level hierarchy with one FCG per app and one CFG per function.}
  \label{fig:higraph_overview}
\end{figure}

Our methodology for constructing the \dataset dataset is illustrated in Figure~\ref{fig:higraph_overview}, which combines the construction pipeline (left) with the resulting two-level hierarchical structure (right). The process involves two primary stages: (1) collecting and curating a large-scale, longitudinal dataset of Android applications, and (2) extracting hierarchical graph representations (FCGs and CFGs) from each application.

\subsection{Data Collection and Curation}

We collected 499,981 Android applications from AndroZoo~\cite{allix2016androzoo}, a repository of apps spanning from 2012 to December 2022. We established ground truth labels by analyzing VirusTotal~\cite{virustotal2025} reports, obtained via their academic API. An application is labeled as malicious if detected by at least 15 antivirus engines, a common threshold in malware research \cite{zhang2020enhancing}. Benign samples are those with no detections. This process resulted in 50,661 malicious and 449,320 benign applications. For malware samples, we further use AVClass2~\cite{sebastian2020avclass2} to assign a fine-grained family label. Aggregate and per-family graph statistics are deferred to Table~\ref{tab:dataset_stats} in Appendix~\ref{sec:dataset_details}, with the per-family characteristics discussed in Section~\ref{sec:malware_classification}.

The dataset construction prioritizes two critical properties. First, \textbf{temporal consistency} is maintained by ensuring samples are distributed evenly across the 10-year collection period. This mitigates concept drift, where malware characteristics evolve over time. We limited our collection to applications up to December 2022 because antivirus detection results continue to evolve over the year following collection~\cite{allix2016androzoo}; the 2012--2020 cohort therefore offers the most stable VirusTotal labels, while 2021--2022 entries should be read alongside their sample budget (Appendix~\ref{sec:limitations}). Second, \textbf{spatial consistency} is achieved by maintaining a realistic malware to benign ratio of approximately 1:9, mirroring real world distributions. This careful curation addresses common dataset biases that can lead to over-optimistic and unrealistic performance evaluations. Table~\ref{tab:yearly_stats} provides a year-by-year breakdown, confirming the consistency of the malware ratio ($\approx$10\%) across all years, and illustrating the significant growth in application complexity over the decade. Note that 6,108 applications ($\approx$1.2\%) lack reliable timestamp metadata in AndroZoo and are therefore excluded from the yearly breakdown, though they remain part of the full dataset.

\begin{table}[t]
  \centering
  \caption{Yearly statistics of \dataset.}
  \label{tab:yearly_stats}
  \small
  \setlength{\tabcolsep}{6pt}
  \begin{tabular}{@{}lrrrrrrr@{}}
    \toprule
    \textbf{Year} & \textbf{\# Apps} & \textbf{\# Malicious} & \textbf{\# Benign} & \textbf{M (\%)} & \textbf{Avg.\ Nodes} & \textbf{Avg.\ Edges} & \textbf{\# Families} \\
    \midrule
    2012  & 53,014  &  5,572 & 47,442  & 10.51 &   183.4 &   294.5 & 108 \\
    2013  & 51,504  &  5,300 & 46,204  & 10.29 &   251.8 &   394.0 & 154 \\
    2014  & 51,814  &  5,109 & 46,705  &  9.86 &   356.1 &   552.1 & 177 \\
    2015  & 55,339  &  5,529 & 49,810  &  9.99 &   462.9 &   746.5 & 222 \\
    2016  & 53,430  &  5,017 & 48,413  &  9.39 &   626.8 & 1,035.1 & 244 \\
    2017  & 54,279  &  5,683 & 48,596  & 10.47 &   574.4 &   954.8 & 232 \\
    2018  & 51,518  &  5,461 & 46,057  & 10.60 &   707.8 & 1,221.4 & 179 \\
    2019  & 53,575  &  5,754 & 47,821  & 10.74 &   938.1 & 1,679.3 & 170 \\
    2020  & 46,589  &  4,673 & 41,916  & 10.03 & 1,131.4 & 2,058.6 & 105 \\
    2021  & 21,986  &  1,866 & 20,120  &  8.48 & 1,461.7 & 2,746.2 &  69 \\
    2022  &    825  &     67 &    758  &  8.12 & 1,677.9 & 3,157.8 &   6 \\
    \midrule
    Total & 493,873 & 50,031 & 443,842 & 10.13 &   616.3 & 1,062.5 & 683 \\
    \bottomrule
  \end{tabular}
\end{table}

\begin{figure}[t]
  \centering
  \begin{subfigure}[b]{0.32\textwidth}
    \centering
    \includegraphics[width=\linewidth]{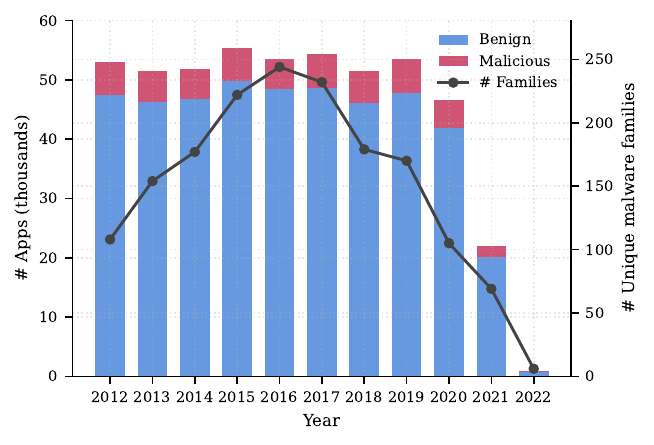}
    \caption{Yearly composition.}
    \label{fig:yearly_trends}
  \end{subfigure}
  \hfill
  \begin{subfigure}[b]{0.32\textwidth}
    \centering
    \includegraphics[width=\linewidth]{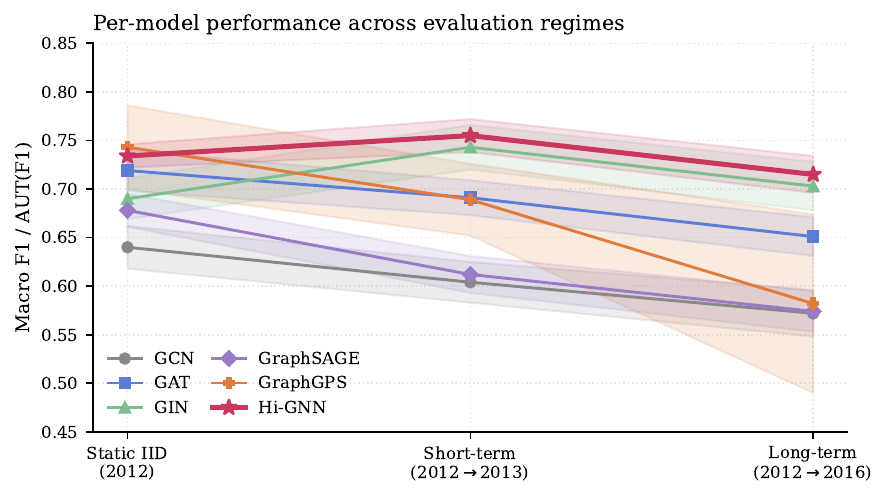}
    \caption{Per-model performance.}
    \label{fig:per_model_aut}
  \end{subfigure}
  \hfill
  \begin{subfigure}[b]{0.32\textwidth}
    \centering
    \includegraphics[width=\linewidth]{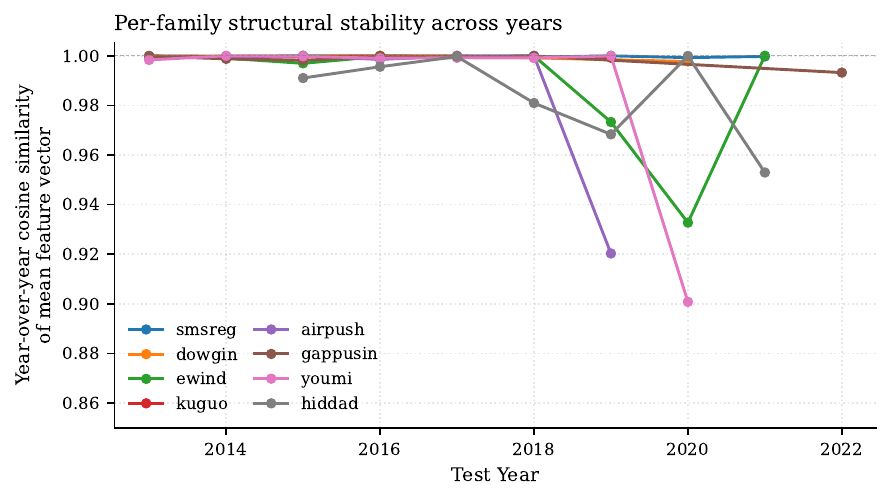}
    \caption{Per-family stability.}
    \label{fig:family_stability}
  \end{subfigure}
  \caption{Three temporal views of \dataset. (a) Yearly app counts (stacked bars: benign/malicious) and unique malware families (line). (b) Macro F1 / AUT(F1) for all six detectors across IID, short-term, and long-term regimes; bands are mean$\pm$std. (c) Year-over-year cosine similarity of mean feature vectors for the eight most prevalent malware families.}
  \label{fig:temporal_overview}
\end{figure}

\subsection{Hierarchical Graph Extraction}
\label{sec:graph_extraction}

After curating the dataset, we use Androguard~\cite{desnos2018androguard} to decompile each application and extract its program structure as a hierarchical graph. This representation consists of a single global Function Call Graph (FCG) per application, with each node in the FCG corresponding to a local function represented by its own Control Flow Graph (CFG). The final \dataset contains over 200 million CFGs and nearly 500,000 FCGs (per-family breakdown in Appendix~\ref{sec:dataset_details}, Table~\ref{tab:dataset_stats}; full min/mean/max statistics also there).

\paragraph{Function Call Graph (FCG).} The FCG provides an inter-procedural view of an application. We model it as a directed graph $\mathcal{G} = (\mathcal{V}, \mathcal{E})$, where nodes $\mathcal{V}$ are functions and edges $\mathcal{E}$ represent calls between them. We distinguish between external functions provided by standard libraries and local functions custom-coded by developers. As malicious logic typically resides in local functions, our analysis focuses on the subgraph induced by these $\mathcal{V}_{loc}$. Furthermore, to reduce noise and focus on security relevant interactions, we apply high-sensitivity filters to retain only call graph edges that connect to sensitive APIs known to be associated with malicious activities.

\paragraph{Control Flow Graph (CFG).} The CFG provides an intra-procedural view of a function's logic. For each local function $f$, we construct its CFG, $G_f = (V_f, E_f)$, where each node $u \in V_f$ is a basic block (a sequence of non-branching instructions), and edges represent control flow transfers. To capture the semantics of each basic block, we engineer an 11-dimensional feature vector for each node, drawing inspiration from prior work~\cite{ling2022malgraph}. This vector summarizes the block from three perspectives: (1) \textbf{Instruction Semantics}, by counting seven categories of bytecode operations (e.g., arithmetic, logic, call); (2) \textbf{Content Metrics}, such as the total instruction count and the presence of constants; and (3) \textbf{Structural Properties}, represented by the block's out-degree. Because these features are derived from bytecode, our representation is independent of the original source language (e.g., Java or Kotlin), making it robust to language evolution.

\subsection{Reproducibility, Accessibility, and Maintenance}
\label{sec:quality_accessibility}

The \dataset and our entire data-processing pipeline (decompilation, graph extraction, feature engineering, all with comprehensive documentation) are publicly released as open-source software at \url{https://higraph.org} under a CC BY-NC-SA 4.0 license, enabling other researchers to replicate our methodology, extend it, and apply consistent processing standards across studies. Alongside direct downloads, the project website provides an interactive interface for graph exploration and statistical summaries, and we maintain a version-controlled release plan with a community feedback mechanism to keep the dataset accurate and current.

\section{Empirical Analysis}
\label{sec:analysis}

This section presents an empirical analysis of our dataset, focusing on the structural properties of Control Flow Graphs (CFGs) and Function Call Graphs (FCGs). We compare the characteristics of malicious and benign applications to uncover patterns that can inform the design and interpretation of graph based detection models. Table~\ref{tab:dataset_comparison} (Appendix~\ref{sec:dataset_details}) compares \dataset with other large scale hierarchical graph datasets: with over 201M local CFGs and 499,981 global FCGs, \dataset vastly exceeds prior datasets in size, and its average of 741.10 nodes per global graph (FCG) far exceeds domains like scientific publications (Arxiv, 30.9) or social networks (QQ, 291.2), while per-CFG (local) basic-block transitions remain modest (avg.\ 12.17 edges), reflecting the inherent sparsity of software control flow.

\subsection{Malware Classification and Family Analysis}
\label{sec:malware_classification}

To provide fine grained malware categorization, we employ AVClass2~\cite{sebastian2020avclass2} to generate standardized malware family labels from VirusTotal detection names. This process leverages the comprehensive taxonomy of AVClass2 to assign structured labels categorized as Family (FAM), Behavior (BEH), Class (CLASS), and File Properties (FILE). This systematic approach ensures robust labeling and enables detailed analysis of malware characteristics across different categories.

Table~\ref{tab:dataset_stats} (Appendix~\ref{sec:dataset_details}) reports the distribution and graph statistics across the most prevalent malware classes alongside benign applications. The analysis reveals distinct structural patterns: \textit{Grayware} represents the largest category with 27,018 samples, typically exhibiting moderate complexity with an average of 141.76 nodes per FCG. In contrast, \textit{Adware} samples, while fewer in number (25,633), display significantly higher structural complexity with an average of 337.38 nodes per FCG and correspondingly higher edge counts (615.80 edges). This structural divergence reflects the different operational requirements of these malware types. Adware often requires more complex interaction mechanisms for advertisement delivery and user engagement tracking.

Notably, \textit{Downloader} malware exhibits the highest structural complexity, with an average of 1,453.33 nodes and 3,370.85 edges per FCG, despite having only 159 samples. This elevated complexity reflects the sophisticated coordination required for payload delivery, network communication, and evasion techniques. The detailed breakdown of structural properties across malware classes (complete min/mean/max statistics available in Appendix~\ref{sec:dataset_details}) provides valuable insights for developing class specific detection strategies and understanding the operational characteristics of different malware families.
\subsection{Structural Comparison of Malicious and Benign Graphs}
\label{sec:structural_comparison}

\begin{figure*}[t]
  \centering
  \begin{subfigure}[b]{0.24\textwidth}
    \centering
    \includegraphics[width=\textwidth]{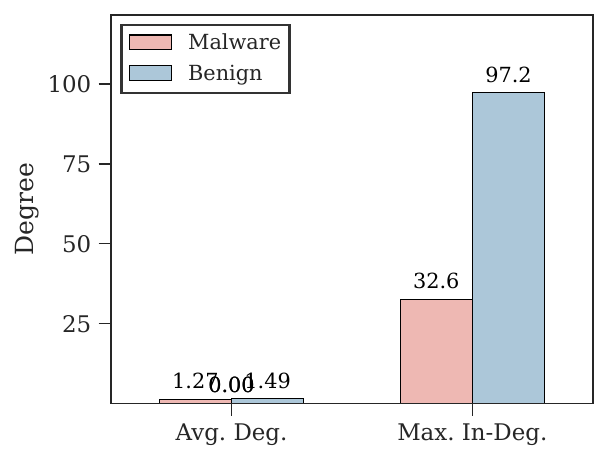}
    \caption{FCG Degree}
    \label{fig:fcg_degree}
  \end{subfigure}%
  \begin{subfigure}[b]{0.24\textwidth}
    \centering
    \includegraphics[width=\textwidth]{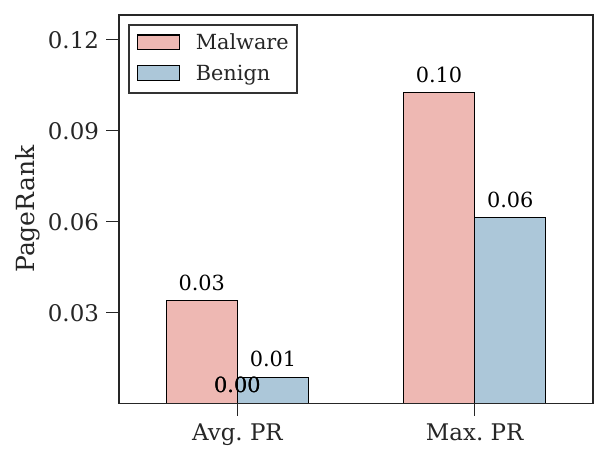}
    \caption{FCG PageRank}
    \label{fig:fcg_pagerank}
  \end{subfigure}%
  \begin{subfigure}[b]{0.24\textwidth}
    \centering
    \includegraphics[width=\textwidth]{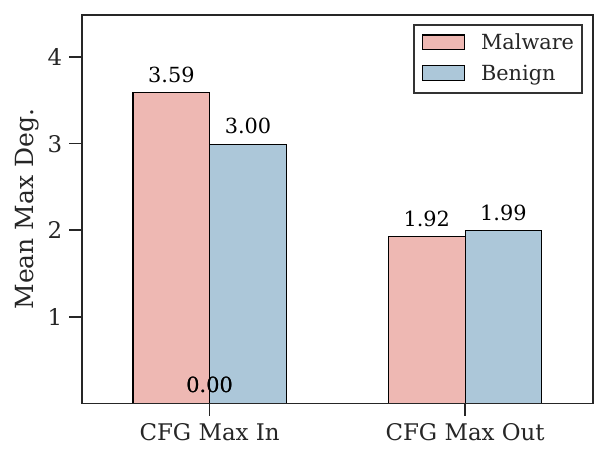}
    \caption{CFG Degree}
    \label{fig:cfg_degree}
  \end{subfigure}%
  \begin{subfigure}[b]{0.24\textwidth}
    \centering
    \includegraphics[width=\textwidth]{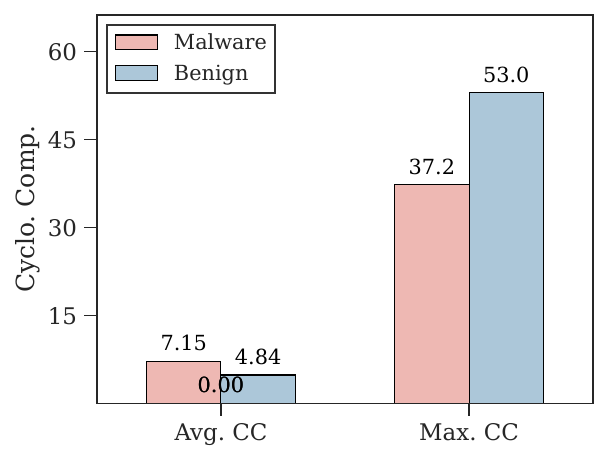}
    \caption{CFG Complexity}
    \label{fig:cfg_complexity}
  \end{subfigure}\\[2pt]
  \begin{subfigure}[b]{0.24\textwidth}
    \centering
    \includegraphics[width=\textwidth]{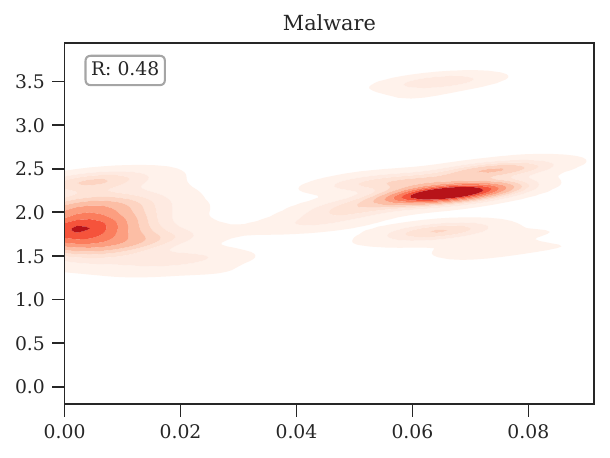}
    \caption{Malware ($R{=}0.48$)}
    \label{fig:cfg_correlation}
  \end{subfigure}%
  \begin{subfigure}[b]{0.24\textwidth}
    \centering
    \includegraphics[width=\textwidth]{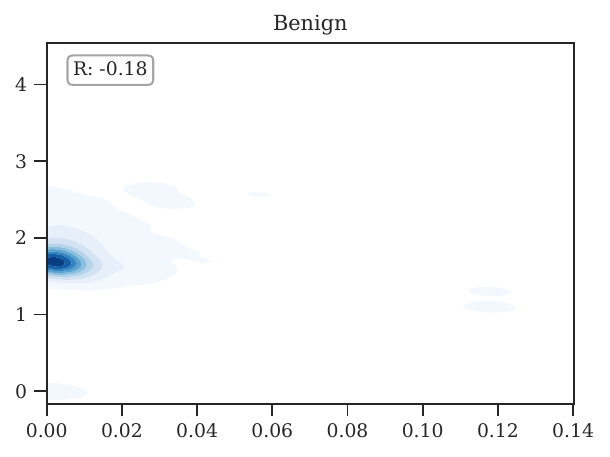}
    \caption{Benign ($R{=}{-}0.18$)}
    \label{fig:fcg_correlation}
  \end{subfigure}%
  \begin{subfigure}[b]{0.24\textwidth}
    \centering
    \includegraphics[width=\textwidth]{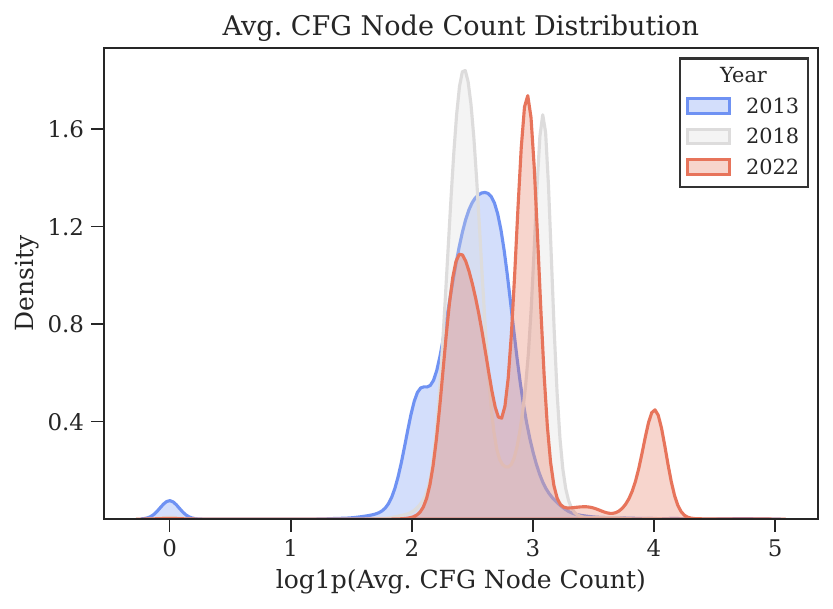}
    \caption{CFG Node Dist.}
    \label{fig:cfg_node_count}
  \end{subfigure}%
  \begin{subfigure}[b]{0.24\textwidth}
    \centering
    \includegraphics[width=\textwidth]{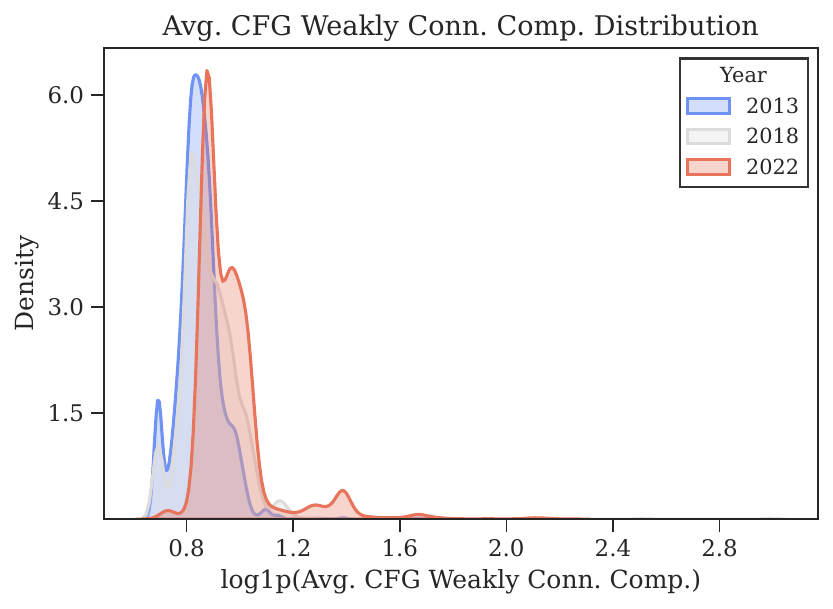}
    \caption{CFG WCC Dist.}
    \label{fig:cfg_num_weakly_connected_components}
  \end{subfigure}
  \caption{Structural comparison of malware vs.\ benign in \dataset. Top (a--d): Avg.\ vs.\ Max for FCG/CFG degree, PageRank, and cyclomatic complexity. Bottom (e--h): 2D density of FCG PageRank vs.\ CFG complexity, and yearly KDEs of CFG node count and WCC count.}
  \label{fig:structural_comparisons_overall}
\end{figure*}

Our analysis of structural metrics, depicted in Figure~\ref{fig:structural_comparisons_overall}, reveals distinct patterns that differentiate malicious and benign applications across both summary statistics (top row) and joint distributions (bottom row).

On the Function Call Graph (FCG) level, malware exhibits higher average and maximum PageRank values (Figure~\ref{fig:structural_comparisons_overall}b). This indicates that malicious applications feature more influential functions and a more centralized overall architecture, where certain functions act as critical hubs for control flow or obfuscation.

On the Control Flow Graph (CFG) level, malware samples show higher node degrees (Figure~\ref{fig:structural_comparisons_overall}c) and elevated cyclomatic complexity (Figure~\ref{fig:structural_comparisons_overall}d). These characteristics point to more intricate conditional logic within individual functions. The presence of exceptionally high maximum values for these metrics suggests that malicious code not only is more complex on average but also contains specific, highly convoluted functions; investigating whether these correspond to obfuscation or to sophisticated payload logic is left to future work.

\paragraph{Joint distribution and correlation.} The joint distribution of FCG PageRank vs.\ CFG cyclomatic complexity (Figure~\ref{fig:structural_comparisons_overall}e--f) reveals contrasting patterns: malware exhibits a moderate positive correlation ($R{=}0.48$) where the most central functions are also the most convoluted, while benign software shows a weak negative correlation ($R{=}{-}0.18$). The yearly KDE distributions (panels g--h) further confirm that benign apps drift toward larger, more fragmented internal structures while malware retains compact, tightly connected logic.

\paragraph{Statistical evidence for hierarchical localization.} To quantify where the malware-vs-benign signal lives in the hierarchy, we run Welch's $t$-tests on per-app aggregate features across all apps with non-empty extracted CFGs (50,252 malware vs.\ 442,225 benign; see Table~\ref{tab:stat_significance} caption for the small discrepancy from Table~\ref{tab:yearly_stats}), reporting Cohen's $d$ effect sizes. Only \emph{CFG-level cyclomatic complexity} reaches a medium effect ($d{=}{+}0.48$, $p{<}10^{-3}$); FCG-level aggregate metrics show negligible effect even though their $p$-values are tiny by virtue of the large sample size. This finding is the empirical foundation for hierarchical modeling: the discriminative signal is concentrated at the CFG level and propagates upward only through learned aggregation, which is precisely what \dataset enables and flat-graph datasets cannot.

\begin{table}[t]
  \centering
  \caption{Statistical evidence for malware-vs-benign structural separation in \dataset, computed on the subset of apps with non-empty Androguard-extracted CFGs (50,252 malware vs.\ 442,225 benign; differences from Table~\ref{tab:yearly_stats} reflect apps whose decompilation produced no extractable CFG and are therefore excluded from CFG-level tests). Welch's $t$-test with Cohen's $d$ effect size; $|d|{>}0.5$ conventionally indicates a medium effect. Only \textbf{CFG cyclomatic complexity} reaches a practically meaningful effect, motivating hierarchical modeling: aggregate FCG-level signals alone are insufficient.}
  \label{tab:stat_significance}
  \small
  \setlength{\tabcolsep}{8pt}
  \begin{tabular}{@{}llrrl@{}}
    \toprule
    \textbf{Level} & \textbf{Metric} & \textbf{Malware (mean$\pm$std)} & \textbf{Benign (mean$\pm$std)} & \textbf{$p$ \;\;\, Cohen's $d$} \\
    \midrule
    \multirow{2}{*}{CFG} & Cyclomatic complexity & $5.20 \pm 2.67$ & $4.09 \pm 1.92$ & $<10^{-3}$ \;\; \cellcolor{red!12}$\bm{+0.48}$ \\
                        & Max degree            & $1.56 \pm 0.21$ & $1.57 \pm 0.27$ & $<10^{-56}$\;\; $-0.07$ \\
    \midrule
    \multirow{2}{*}{FCG} & Avg.\ in-degree       & $1.47 \pm 0.47$ & $1.48 \pm 0.32$ & $0.013$ \;\;\;\, $-0.01$ \\
                        & Max degree            & --              & --             & not significant \\
    \bottomrule
  \end{tabular}
\end{table}

Collectively, these findings show that the discriminative signal between malicious and benign software is primarily concentrated at the intra-procedural (CFG) level---most notably in cyclomatic complexity ($d{=}{+}0.48$)---with secondary inter-procedural (FCG) patterns visible in joint distributions and yearly KDEs but not in single-summary aggregates; this divergence persists across temporal evolution, providing a structurally grounded signal that motivates hierarchy-aware modeling.

\section{Evaluating \dataset}
\label{sec:application}

We evaluate \dataset on core malware analysis tasks to demonstrate its effectiveness. Our central hypothesis is that \dataset's hierarchical structure, which combines intraprocedural Control Flow Graphs (CFGs) with an interprocedural Function Call Graph (FCG), captures rich program semantics. This structure, we argue, leads to more robust and accurate malware detection. Specifically, we seek to answer the following research questions:

\begin{itemize}[topsep=1pt, itemsep=1pt, parsep=0pt, leftmargin=*]
  \item \textbf{RQ1 (Efficacy):} Can graph learning models effectively leverage \dataset for malware detection and classification?
  \item \textbf{RQ2 (Hierarchy):} Does the hierarchical structure yield superior performance compared to single-level graph representations?
  \item \textbf{RQ3 (Robustness):} How resilient are hierarchical representations to temporal concept drift?
\end{itemize}

\subsection{Malware Detection and Classification (RQ1 \& RQ2)}
We benchmark \dataset on Binary Detection (Malware vs. Benign); we additionally provide family labels and report descriptive family-level statistics (top-5/top-10/top-20 and the full 683-family taxonomy, per-family statistics in Appendix~\ref{sec:dataset_details}) for use in downstream classification studies.

\paragraph{Models.} As single-level FCG baselines we use GCN~\cite{kipf2017semi}, GAT~\cite{velivckovic2018graph}, GIN~\cite{xu2018powerful}, and GraphSAGE~\cite{hamilton2017inductive}. To represent recent transformer-based architectures we also evaluate GraphGPS~\cite{rampavsek2022recipe}, which combines local GIN message passing with global multi-head self-attention. To exploit \dataset's hierarchy we implement Hi-GNN, a minimal model with separate two-layer GCN encoders for CFGs and FCGs whose multi-level representations are fused before the final classifier. All models share the same 11-dim bytecode input features (instruction counts, constants, etc.) and a unified 128-dim hidden representation; single-level baselines aggregate per-function CFG features by mean pooling, with nodes lacking CFGs (e.g.\ external APIs) assigned in/out-degree as structural features. Any performance gap is therefore attributable to architectural and hierarchical modeling choices rather than feature disparity. \dataset additionally provides raw function names and bytecode sequences for use with pre-trained code models~\cite{guo2020graphcodebert}.

\paragraph{Setup.} We evaluate detectors across the full 2012--2022 range under three regimes: (i) an IID regime on the 2012 cohort (53{,}014 apps, 70/15/15 split) so any drop quantifies temporal degradation rather than within-year generalization; (ii) a short-term regime trained on 2012 and tested month-by-month over 2013; and (iii) a long-horizon regime extending through 2016, plus a Train 2016 anchor covering 2017--2022 via the GraphGPS protocol. Following standard concept-drift evaluation~\cite{zhang2020enhancing,pendlebury2019tesseract}, models are trained with Adam (Appendix~\ref{sec:experimental_setup}); we report PR-AUC and Macro F1 (headline metric given $\sim$10\% class imbalance), averaged over three random seeds.

\paragraph{Results.} Table~\ref{tab:benchmark_combined} consolidates static IID detection (RQ1, RQ2) with longitudinal concept-drift robustness (RQ3) for all six models. On the static 2012 IID split, all GNNs achieve reasonable Macro F1 ($0.640$--$0.761$) and PR-AUC ($0.615$--$0.785$), confirming \dataset's learnability; Hi-GNN attains the best static IID Macro F1 ($0.761$) and PR-AUC ($0.785$), with GraphGPS a close second on both metrics. Under temporal drift, summarized by $\mathrm{AUT}(f,N){=}\tfrac{1}{N-1}\sum_{k=0}^{N-1}\tfrac{f(k+1)+f(k)}{2}$~\cite{zhang2020enhancing} where $f$ is the metric and $N$ is the number of test months, Hi-GNN sustains the strongest performance: AUT(F1) of $0.755$ for 2012$\rightarrow$2013 (vs.\ $0.604$--$0.743$ for single-level GNNs and $0.689$ for GraphGPS) and $0.715$ for the longer 2012$\rightarrow$2016 horizon. On AUT(PR-AUC), GraphGPS leads at the short horizon ($0.673$ vs.\ Hi-GNN's $0.564$) but falls behind at the long horizon ($0.462$ vs.\ $0.498$), consistent with richer learned aggregation helping at short range while being more sensitive to cumulative drift; we therefore report Macro F1 as the headline metric, given the $\sim$10\% class imbalance. Despite its global self-attention mechanism, GraphGPS lags Hi-GNN on AUT(F1) at both horizons under the same 128-dim hidden representation, indicating that, in our setup, hierarchical aggregation provides drift robustness that a flat-graph transformer's added capacity does not match. This sustained robustness answers RQ3: the hierarchy lets Hi-GNN combine stable CFG-level semantics with adaptable FCG-level architecture, identifying persistent behavioral patterns that transcend temporal shifts.

\begin{table}[t]
  \centering
  \caption{Benchmark of graph-based detectors on \dataset across three regimes: static IID (2012, 70/15/15 split) and Area Under Time on 2012$\rightarrow$2013 and 2012$\rightarrow$2016. All models trained with three random seeds on the same 2012 IID split; mean$\pm$std reported across seeds. Best per metric in \textbf{bold}.}
  \label{tab:benchmark_combined}
  \small
  \setlength{\tabcolsep}{4pt}
  \resizebox{\linewidth}{!}{%
  \begin{tabular}{@{}lcccccc@{}}
    \toprule
    & \multicolumn{2}{c}{\textbf{Static IID (2012)}}
    & \multicolumn{2}{c}{\textbf{Short-term: 2012$\rightarrow$2013}}
    & \multicolumn{2}{c}{\textbf{Long-term: 2012$\rightarrow$2016}} \\
    \cmidrule(lr){2-3} \cmidrule(lr){4-5} \cmidrule(lr){6-7}
    \textbf{Model} & \textbf{PR-AUC} & \textbf{Macro F1}
                   & \textbf{AUT(PR-AUC)} & \textbf{AUT(F1)}
                   & \textbf{AUT(PR-AUC)} & \textbf{AUT(F1)} \\
    \midrule
    GCN          & $0.627\!\pm\!0.025$ & $0.640\!\pm\!0.022$ & $0.516\!\pm\!0.015$ & $0.604\!\pm\!0.021$ & $0.489\!\pm\!0.018$ & $0.572\!\pm\!0.024$ \\
    GAT          & $0.641\!\pm\!0.012$ & $0.719\!\pm\!0.021$ & $0.513\!\pm\!0.012$ & $0.691\!\pm\!0.018$ & $0.483\!\pm\!0.014$ & $0.651\!\pm\!0.020$ \\
    GIN          & $0.615\!\pm\!0.035$ & $0.690\!\pm\!0.021$ & $0.550\!\pm\!0.019$ & $0.743\!\pm\!0.023$ & $\mathbf{0.520\!\pm\!0.021}$ & $0.703\!\pm\!0.025$ \\
    GraphSAGE    & $0.619\!\pm\!0.014$ & $0.678\!\pm\!0.017$ & $0.518\!\pm\!0.014$ & $0.612\!\pm\!0.019$ & $0.489\!\pm\!0.016$ & $0.574\!\pm\!0.021$ \\
    GraphGPS     & $0.774\!\pm\!0.064$ & $0.743\!\pm\!0.043$ & $\mathbf{0.673\!\pm\!0.100}$ & $0.689\!\pm\!0.037$ & $0.462\!\pm\!0.192$ & $0.582\!\pm\!0.092$ \\
    \midrule
    Hi-GNN       & $\mathbf{0.785\!\pm\!0.012}$ & $\mathbf{0.761\!\pm\!0.028}$ & $0.564\!\pm\!0.011$ & $\mathbf{0.755\!\pm\!0.017}$ & $0.498\!\pm\!0.026$ & $\mathbf{0.715\!\pm\!0.019}$ \\
    \bottomrule
  \end{tabular}%
  }
\end{table}

\subsection{Temporal Robustness to Malware Evolution (RQ3)}
\label{sec:temporal}
\begin{figure}[t]
  \centering
  \includegraphics[width=\linewidth]{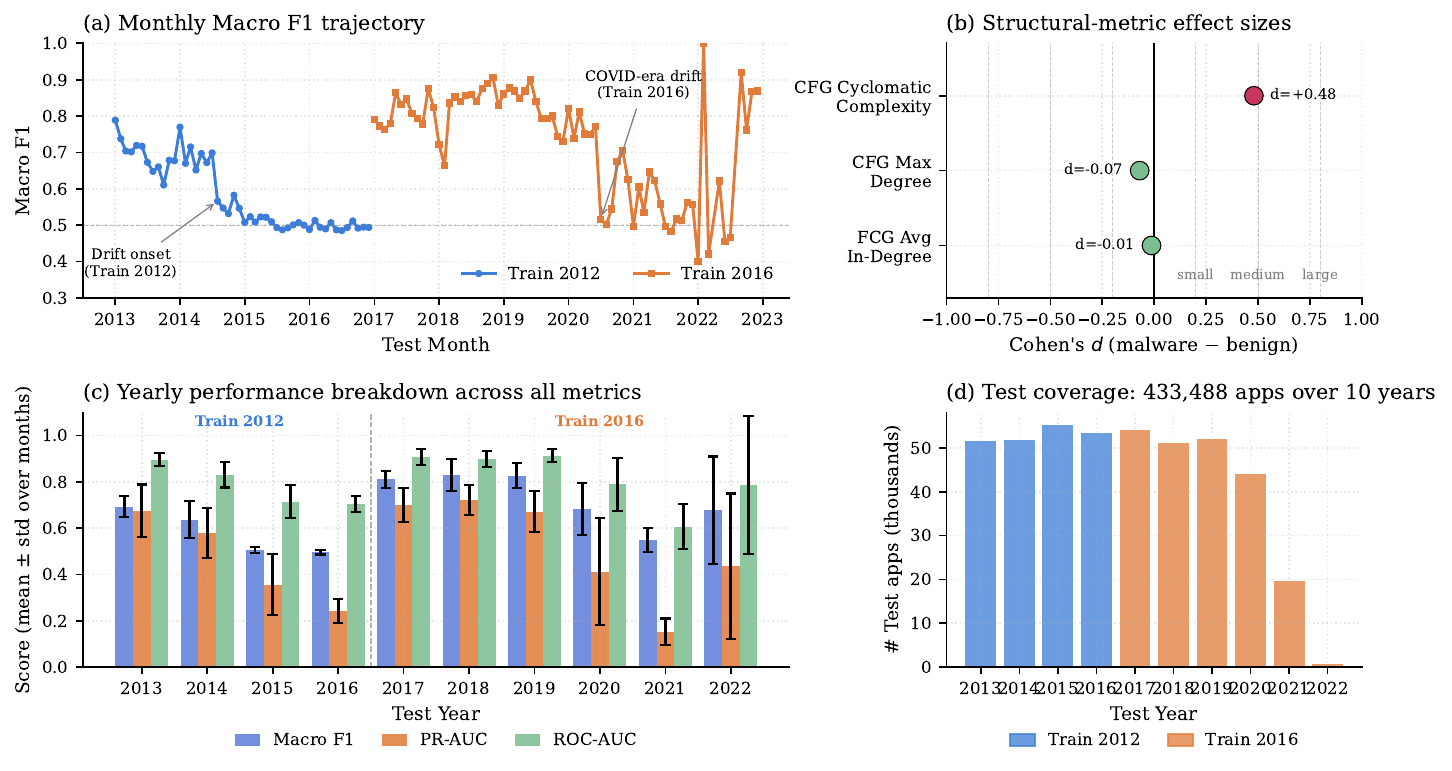}
  \caption{Concept-drift analysis of GraphGPS on \dataset under two anchors. (a) Monthly F1 trajectory; (b) malware-vs-benign effect sizes; (c) yearly metric breakdown; (d) test-app coverage (433K over 10 years).}
  \label{fig:drift_dashboard}
\end{figure}

To probe drift beyond the 2012-anchored AUT, we evaluate GraphGPS under two training anchors (Train 2012 and Train 2016), covering 118 monthly evaluation points and \textbf{433{,}488} test apps ($\approx$87\% of \dataset). Figure~\ref{fig:drift_dashboard} visualizes the monthly trajectory; Table~\ref{tab:graphgps_yearly} reports the yearly mean$\pm$std for Macro F1, PR-AUC, and ROC-AUC across both anchors.

Three observations stand out. (i) The monthly trajectories (Figure~\ref{fig:drift_dashboard}a) reveal sharp drift events: Train 2012 drops from $0.69$ in 2013 to $0.50$ by 2015, while Train 2016 holds $\sim$0.83 through 2019 before a 2020-Q3 inflection. (ii) Structural separability between malware and benign (Figure~\ref{fig:drift_dashboard}b) concentrates at the CFG level---only cyclomatic complexity reaches a medium Cohen's $d$ of $+0.48$, providing direct empirical support for hierarchical modeling. (iii) The Train 2016 anchor shows a much higher overall AUT(F1) of $0.730$ across 70 months than the Train 2012 anchor's $0.582$ across 48 months, indicating that the 2016--2019 family ecosystem is structurally more stable than the 2012--2015 era of rapid family churn.

\begin{table}[t]
  \centering
  \caption{Year-by-year GraphGPS performance on \dataset under two temporal anchors, with monthly mean$\pm$std aggregated within each test year. Test apps total \textbf{433{,}488} (Train 2012: 211{,}919 over 48 months; Train 2016: 221{,}569 over 70 months), spanning 10 test years and $\approx$87\% of \dataset. The graceful but non-monotonic degradation indicates real, dataset-driven concept drift rather than fitting noise.}
  \label{tab:graphgps_yearly}
  \small
  \setlength{\tabcolsep}{4pt}
  \resizebox{\linewidth}{!}{%
  \begin{tabular}{@{}llcccccccccc@{}}
    \toprule
    \textbf{Anchor} & \textbf{Metric} & \textbf{2013} & \textbf{2014} & \textbf{2015} & \textbf{2016} & \textbf{2017} & \textbf{2018} & \textbf{2019} & \textbf{2020} & \textbf{2021} & \textbf{2022} \\
    \midrule
    \multirow{4}{*}{Train 2012}
        & Macro F1   & $0.69\!\pm\!.05$ & $0.64\!\pm\!.08$ & $0.51\!\pm\!.01$ & $0.50\!\pm\!.01$ & --                   & --                   & --                   & --                   & --                   & -- \\
        & PR-AUC     & $0.68\!\pm\!.11$ & $0.58\!\pm\!.11$ & $0.36\!\pm\!.13$ & $0.24\!\pm\!.05$ & --                   & --                   & --                   & --                   & --                   & -- \\
        & ROC-AUC    & $0.90\!\pm\!.03$ & $0.83\!\pm\!.05$ & $0.71\!\pm\!.07$ & $0.71\!\pm\!.03$ & --                   & --                   & --                   & --                   & --                   & -- \\
        & N apps     & 51,496           & 51,814           & 55,323           & 53,286           & --                   & --                   & --                   & --                   & --                   & -- \\
    \midrule
    \multirow{4}{*}{Train 2016}
        & Macro F1   & --               & --               & --               & --               & $0.81\!\pm\!.04$ & $0.83\!\pm\!.07$ & $0.83\!\pm\!.05$ & $0.68\!\pm\!.11$ & $0.55\!\pm\!.05$ & $0.68\!\pm\!.23$ \\
        & PR-AUC     & --               & --               & --               & --               & $0.70\!\pm\!.07$ & $0.72\!\pm\!.07$ & $0.67\!\pm\!.09$ & $0.41\!\pm\!.23$ & $0.15\!\pm\!.06$ & $0.43\!\pm\!.31$ \\
        & ROC-AUC    & --               & --               & --               & --               & $0.91\!\pm\!.04$ & $0.90\!\pm\!.03$ & $0.91\!\pm\!.03$ & $0.79\!\pm\!.11$ & $0.61\!\pm\!.10$ & $0.79\!\pm\!.30$ \\
        & N apps     & --               & --               & --               & --               & 54,130           & 51,012           & 52,014           & 44,100           & 19,615           & 698 \\
    \bottomrule
  \end{tabular}%
  }
\end{table}

\paragraph{Family-level stability.} The earlier per-family analysis (Figure~\ref{fig:temporal_overview}c) decomposes drift to the family level: most families (\textit{smsreg}, \textit{dowgin}, \textit{kuguo}, \textit{gappusin}, \textit{airpush}) exhibit near-perfect year-over-year stability ($\geq$0.99 cosine similarity), suggesting their structural fingerprint is largely preserved. In contrast, \textit{ewind}, \textit{youmi}, and \textit{hiddad} show pronounced dips around 2019--2020, indicating that family-level drift, while concentrated, is not uniformly distributed across the ecosystem. Beyond detector behavior, the underlying graph structures themselves diverge across the decade: benign FCGs grow and become sparser while malware FCGs shrink but densify (Appendix~\ref{sec:extended_temporal}).

\paragraph{Limitations and responsible use.} Five caveats apply: (i) the conservative VT$\geq$15 threshold leaves only 67 confirmed malware in 2022, so 2021--2022 drift metrics should be read alongside their sample budget; (ii) \dataset is Android-only, though the CFG/FCG pipeline is platform-agnostic; (iii) the representation is static (no runtime behavior); (iv) AVClass2 family labels inherit upstream antivirus naming noise; (v) an XGBoost baseline on graph statistics (Appendix~\ref{sec:xgboost_baseline}) matches GraphGPS-level temporal AUT(F1) but falls well short of Hi-GNN; parameter-matched and randomized-hierarchy ablations remain future work (Appendix~\ref{sec:limitations}). \dataset is intended for defensive research: we release only graph-level abstractions (no executable bytecode), and raw APK access remains gated by AndroZoo's academic-use credentials~\cite{allix2016androzoo}.

\section{Conclusion}
\label{sec:conclusion}

We introduced \dataset, a hierarchical graph dataset of 499K Android applications with 200M+ nested CFGs and 499K FCGs spanning 2012--2022, along with a longitudinal benchmark and cross-anchor concept-drift protocol covering 433{,}488 test apps over 10 years. Our experiments show that hierarchical CFG/FCG modeling delivers the strongest static IID accuracy and stronger temporal AUT(F1) than flat-graph GNNs and a modern graph transformer~\cite{rampavsek2022recipe} under matched feature and hidden dimensions, and that the discriminative malware signal localizes to CFG-level cyclomatic complexity rather than aggregate FCG metrics, advancing drift-aware evaluation of graph-based malware detection.

\bibliographystyle{plainnat}
\bibliography{main}

@inproceedings{allix2016androzoo,
  title     = {Androzoo: Collecting millions of android apps for the research community},
  author    = {Allix, Kevin and Bissyand{\'e}, Tegawend{\'e} F and Klein, Jacques and Le Traon, Yves},
  booktitle = {Proceedings of the 13th International Conference on Mining Software Repositories},
  pages     = {468--471},
  year      = {2016},
  doi       = {10.1145/2901739.2903508}
}

@inproceedings{arp2014drebin,
  title     = {Drebin: Effective and explainable detection of android malware in your pocket.},
  author    = {Arp, Daniel and Spreitzenbarth, Michael and Hubner, Malte and Gascon, Hugo and Rieck, Konrad and Siemens, CERT},
  booktitle = {Ndss},
  volume    = {14},
  number    = {1},
  pages     = {23--26},
  year      = {2014}
}

@inproceedings{arp2022dos,
  title     = {Dos and don'ts of machine learning in computer security},
  author    = {Arp, Daniel and Quiring, Erwin and Pendlebury, Feargus and Warnecke, Alexander and Pierazzi, Fabio and Wressnegger, Christian and Cavallaro, Lorenzo and Rieck, Konrad},
  booktitle = {31st USENIX Security Symposium (USENIX Security 22)},
  pages     = {3971--3988},
  year      = {2022}
}

@article{bilotSurveyMalwareDetection2023,
  title   = {A Survey on Malware Detection with Graph Representation Learning},
  author  = {Bilot, Tristan and Madhoun, Nour El and Agha, Khaldoun Al and Zouaoui, Anis},
  journal = {arXiv preprint arXiv:2303.16004},
  year    = {2023},
  doi     = {10.48550/arXiv.2303.16004}
}

@article{chen2023denoising,
  title     = {Denoising variational graph of graphs auto-encoder for predicting structured entity interactions},
  author    = {Chen, Han and Wang, Hanchen and Chen, Hongmei and Zhang, Ying and Zhang, Wenjie and Lin, Xuemin},
  journal   = {IEEE Transactions on Knowledge and Data Engineering},
  volume    = {36},
  number    = {3},
  pages     = {1016--1029},
  year      = {2023},
  publisher = {IEEE}
}

@article{chen2023guided,
  title     = {Guided malware sample analysis based on graph neural networks},
  author    = {Chen, Yi-Hsien and Lin, Si-Chen and Huang, Szu-Chun and Lei, Chin-Laung and Huang, Chun-Ying},
  journal   = {IEEE Transactions on Information Forensics and Security},
  volume    = {18},
  pages     = {4128--4143},
  year      = {2023},
  publisher = {IEEE}
}

@article{desnos2018androguard,
  title   = {Androguard documentation},
  author  = {Desnos, Anthony and Gueguen, Geoffroy},
  journal = {Obtenido de Androguard},
  year    = {2018}
}

@article{Fey/Lenssen/2019,
  title   = {Fast graph representation learning with PyTorch Geometric},
  author  = {Fey, Matthias and Lenssen, Jan Eric},
  journal = {arXiv preprint arXiv:1903.02428},
  year    = {2019}
}

@article{freitas2020large,
  title   = {A large-scale database for graph representation learning},
  author  = {Freitas, Scott and Dong, Yuxiao and Neil, Joshua and Chau, Duen Horng},
  journal = {arXiv preprint arXiv:2011.07682},
  year    = {2020}
}

@article{gao2023hierarchical,
  title     = {Hierarchical graph learning for protein--protein interaction},
  author    = {Gao, Ziqi and Jiang, Chenran and Zhang, Jiawen and Jiang, Xiaosen and Li, Lanqing and Zhao, Peilin and Yang, Huanming and Huang, Yong and Li, Jia},
  journal   = {Nature Communications},
  volume    = {14},
  number    = {1},
  pages     = {1093},
  year      = {2023},
  publisher = {Nature Publishing Group UK London}
}

@inproceedings{grover2016node2vec,
  title     = {node2vec: Scalable feature learning for networks},
  author    = {Grover, Aditya and Leskovec, Jure},
  booktitle = {Proceedings of the 22nd ACM SIGKDD international conference on Knowledge discovery and data mining},
  pages     = {855--864},
  year      = {2016}
}

@article{guo2020graphcodebert,
  title   = {Graphcodebert: Pre-training code representations with data flow},
  author  = {Guo, Daya and Ren, Shuo and Lu, Shuai and Feng, Zhangyin and Tang, Duyu and Liu, Shujie and Zhou, Long and Duan, Nan and Svyatkovskiy, Alexey and Fu, Shengyu and others},
  journal = {arXiv preprint arXiv:2009.08366},
  year    = {2020}
}

@article{hamilton2017inductive,
  title   = {Inductive representation learning on large graphs},
  author  = {Hamilton, Will and Ying, Zhitao and Leskovec, Jure},
  journal = {Advances in neural information processing systems},
  volume  = {30},
  year    = {2017}
}

@article{he2022msdroid,
  title     = {Msdroid: Identifying malicious snippets for android malware detection},
  author    = {He, Yiling and Liu, Yiping and Wu, Lei and Yang, Ziqi and Ren, Kui and Qin, Zhan},
  journal   = {IEEE Transactions on Dependable and Secure Computing},
  volume    = {20},
  number    = {3},
  pages     = {2025--2039},
  year      = {2022},
  publisher = {IEEE},
  doi       = {10.1109/tdsc.2022.3168285}
}

@article{kipf2017semi,
  title     = {Semi-supervised classification with graph convolutional networks},
  author    = {Kipf, Thomas N and Welling, Max},
  booktitle = {International Conference on Learning Representations (ICLR)},
  year      = {2017}
}

@inproceedings{li2019semi,
  title     = {Semi-supervised graph classification: A hierarchical graph perspective},
  author    = {Li, Jia and Rong, Yu and Cheng, Hong and Meng, Helen and Huang, Wenbing and Huang, Junzhou},
  booktitle = {The World Wide Web Conference},
  pages     = {972--982},
  year      = {2019}
}

@inproceedings{ling2022malgraph,
  title        = {MalGraph: Hierarchical graph neural networks for robust windows malware detection},
  author       = {Ling, Xiang and Wu, Lingfei and Deng, Wei and Qu, Zhenqing and Zhang, Jiangyu and Zhang, Sheng and Ma, Tengfei and Wang, Bin and Wu, Chunming and Ji, Shouling},
  booktitle    = {IEEE INFOCOM 2022-IEEE Conference on Computer Communications},
  pages        = {1998--2007},
  year         = {2022},
  organization = {IEEE}
}

@inproceedings{loGraphNeuralNetworkbased2022,
  title     = {Graph Neural Network-based Android Malware Classification with Jumping Knowledge},
  booktitle = {2022 IEEE Conference on Dependable and Secure Computing (DSC)},
  author    = {Lo, Wai Weng and Layeghy, Siamak and Sarhan, Mohanad and Gallagher, Marcus and Portmann, Marius},
  year      = {2022},
  pages     = {1--9},
  doi       = {10.1109/DSC54232.2022.9888878}
}

@inproceedings{mahdavifar2020dynamic,
  title        = {Dynamic android malware category classification using semi-supervised deep learning},
  author       = {Mahdavifar, Samaneh and Kadir, Andi Fitriah Abdul and Fatemi, Rasool and Alhadidi, Dima and Ghorbani, Ali A},
  booktitle    = {2020 IEEE Intl Conf on Dependable, Autonomic and Secure Computing, Intl Conf on Pervasive Intelligence and Computing, Intl Conf on Cloud and Big Data Computing, Intl Conf on Cyber Science and Technology Congress (DASC/PiCom/CBDCom/CyberSciTech)},
  pages        = {515--522},
  year         = {2020},
  organization = {IEEE}
}

@inproceedings{pendlebury2019tesseract,
  title     = {$\{$TESSERACT$\}$: Eliminating experimental bias in malware classification across space and time},
  author    = {Pendlebury, Feargus and Pierazzi, Fabio and Jordaney, Roberto and Kinder, Johannes and Cavallaro, Lorenzo},
  booktitle = {28th USENIX security symposium (USENIX Security 19)},
  pages     = {729--746},
  year      = {2019}
}

@inproceedings{perozzi2014deepwalk,
  title     = {DeepWalk: Online learning of social representations},
  author    = {Perozzi, Bryan and Al-Rfou, Rami and Skiena, Steven},
  booktitle = {Proceedings of the 20th ACM SIGKDD international conference on Knowledge discovery and data mining},
  pages     = {701--710},
  year      = {2014}
}

@article{rampavsek2022recipe,
  title   = {Recipe for a general, powerful, scalable graph transformer},
  author  = {Ramp{\'a}{\v{s}}ek, Ladislav and Galkin, Michael and Dwivedi, Vijay Prakash and Luu, Anh Tuan and Wolf, Guy and Beaini, Dominique},
  journal = {Advances in Neural Information Processing Systems},
  volume  = {35},
  pages   = {14501--14515},
  year    = {2022}
}

@article{rong2020self,
  title   = {Self-supervised graph transformer on large-scale molecular data},
  author  = {Rong, Yu and Bian, Yatao and Xu, Tingyang and Xie, Weiyang and Wei, Ying and Huang, Wenbing and Huang, Junzhou},
  journal = {Advances in neural information processing systems},
  volume  = {33},
  pages   = {12559--12571},
  year    = {2020}
}

@inproceedings{sebastian2020avclass2,
  title     = {Avclass2: Massive malware tag extraction from av labels},
  author    = {Sebasti{\'a}n, Silvia and Caballero, Juan},
  booktitle = {Proceedings of the 36th Annual Computer Security Applications Conference},
  pages     = {42--53},
  year      = {2020}
}

@article{velivckovic2018graph,
  title     = {Graph attention networks},
  author    = {Veli{\v{c}}kovi{\'c}, Petar and Cucurull, Guillem and Casanova, Arantxa and Romero, Adriana and Lio, Pietro and Bengio, Yoshua},
  booktitle = {International Conference on Learning Representations (ICLR)},
  year      = {2018}
}

@misc{virusshare2025,
  title        = {VirusShare.com},
  author       = {VirusShare},
  howpublished = {https://virusshare.com/},
  note         = {[Accessed on 05/14/2025]},
  year         = {2025}
}

@misc{virustotal2025,
  title        = {VirusTotal},
  author       = {{VirusTotal}},
  howpublished = {\url{https://www.virustotal.com/}},
  note         = {Accessed: 2025-05-14},
  year         = {2025}
}

@article{wang2020gognn,
  title   = {Gognn: Graph of graphs neural network for predicting structured entity interactions},
  author  = {Wang, Hanchen and Lian, Defu and Zhang, Ying and Qin, Lu and Lin, Xuemin},
  journal = {arXiv preprint arXiv:2005.05537},
  year    = {2020}
}

@inproceedings{wang2021multi,
  title     = {Multi-view graph contrastive representation learning for drug-drug interaction prediction},
  author    = {Wang, Yingheng and Min, Yaosen and Chen, Xin and Wu, Ji},
  booktitle = {Proceedings of the web conference 2021},
  pages     = {2921--2933},
  year      = {2021}
}

@inproceedings{wei2017deep,
  title        = {Deep ground truth analysis of current android malware},
  author       = {Wei, Fengguo and Li, Yuping and Roy, Sankardas and Ou, Xinming and Zhou, Wu},
  booktitle    = {Detection of Intrusions and Malware, and Vulnerability Assessment: 14th International Conference, DIMVA 2017, Bonn, Germany, July 6-7, 2017, Proceedings 14},
  pages        = {252--276},
  year         = {2017},
  organization = {Springer}
}

@article{xu2018powerful,
  title   = {How powerful are graph neural networks?},
  author  = {Xu, Keyulu and Hu, Weihua and Leskovec, Jure and Jegelka, Stefanie},
  journal = {arXiv preprint arXiv:1810.00826},
  year    = {2018}
}

@inproceedings{yangCADEDetectingExplaining2021,
  title     = {{CADE}: Detecting and Explaining Concept Drift Samples for Security Applications},
  booktitle = {30th USENIX Security Symposium (USENIX Security 21)},
  author    = {Yang, Limin and Guo, Wenbo and Hao, Qingying and Ciptadi, Arridhana and Ahmadzadeh, Ali and Xing, Xinyu and Wang, Gang},
  year      = {2021},
  pages     = {2327--2344}
}

@inproceedings{zhang2020enhancing,
  title     = {Enhancing State-of-the-art Classifiers with API Semantics to Detect Evolved Android Malware},
  author    = {Zhang, Xiaohan and Zhang, Yuan and Zhong, Ming and Ding, Daizong and Cao, Yinzhi and Zhang, Yukun and Zhang, Mi and Yang, Min},
  booktitle = {Proceedings of the 2020 ACM SIGSAC Conference on Computer and Communications Security},
  series    = {CCS '20},
  pages     = {757--770},
  year      = {2020},
  publisher = {Association for Computing Machinery},
  doi       = {10.1145/3372297.3417291}
}

@misc{zhangSemanticpreservingReinforcementLearning2022,
  title  = {Semantic-Preserving Reinforcement Learning Attack Against Graph Neural Networks for Malware Detection},
  author = {Zhang, Lan and Liu, Peng and Choi, Yoon-Ho and Chen, Ping},
  year   = {2022},
  note   = {arXiv preprint arXiv:2009.05602}
}

@inproceedings{zhangSemanticsAwareAndroidMalware2014,
  title     = {Semantics-Aware Android Malware Classification Using Weighted Contextual API Dependency Graphs},
  booktitle = {Proceedings of the 2014 ACM SIGSAC Conference on Computer and Communications Security},
  author    = {Zhang, Mu and Duan, Yue and Yin, Heng and Zhao, Zhiruo},
  year      = {2014},
  series    = {CCS '14},
  pages     = {1105--1116},
  publisher = {Association for Computing Machinery},
  doi       = {10.1145/2660267.2660359}
}

\newpage
\appendix
\section{Extended Dataset Statistics}
\label{sec:dataset_details}

This appendix provides comprehensive statistical details that supplement the malware classification analysis presented in Section~\ref{sec:malware_classification} of the main paper. The complete breakdowns presented here enable detailed analysis of structural patterns across all identified malware classes and support reproducible research.

\begin{table}[t]
  \centering
  \caption{Comparison of \dataset with other benchmark hierarchical and two-level graph datasets. \dataset is unique in its scale, particularly in the number of individual local graphs (CFGs), and is the only resource targeting cybersecurity at this granularity.}
  \label{tab:comparison}
  \small
  \setlength{\tabcolsep}{6pt}
  \resizebox{\linewidth}{!}{%
  \begin{tabular}{@{}llrrrr@{}}
    \toprule
    \multirow{2}{*}{\textbf{Dataset}}    & \multirow{2}{*}{\textbf{Field}} & \multicolumn{2}{c}{\textbf{Global Graph}} & \multicolumn{2}{c}{\textbf{Local Graph}} \\
    \cmidrule(lr){3-4} \cmidrule(lr){5-6}
                                         &                                 & \textbf{\# Graphs} & \textbf{Avg.\ Nodes} & \textbf{\# Graphs}        & \textbf{Avg.\ Nodes} \\
    \midrule
    CCI900~\cite{chen2023denoising}      & Chemical                        &                  1 &                 25.4 &                    14,343 &                 25.4 \\
    CCI950~\cite{chen2023denoising}      & Chemical                        &                  1 &                 26.2 &                     7,606 &                 26.2 \\
    NetBasedDDI~\cite{chen2023denoising} & Drug                            &                  1 &                 24.8 &                       596 &                 24.8 \\
    ZhangDDI~\cite{chen2023denoising}    & Drug                            &                  1 &                 25.2 &                       544 &                 25.2 \\
    ChChMiner~\cite{chen2023denoising}   & Drug                            &                  1 &                 27.8 &                     1,329 &                 27.8 \\
    DeepDDI~\cite{chen2023denoising}     & Drug                            &                  1 &                 27.5 &                     1,704 &                 27.5 \\
    Arxiv~\cite{li2019semi}              & Text                            &                  1 &                 30.9 &                     4,666 &                 30.9 \\
    QQ~\cite{li2019semi}                 & Social                          &                  1 &                291.2 &                    37,836 &                291.2 \\
    \midrule
    \textbf{\dataset}                    & \textbf{Cybersecurity}          &  \textbf{499,981} &      \textbf{741.1} & \textbf{201,792,085}       &      \textbf{12.2} \\
    \bottomrule
  \end{tabular}%
  }
\end{table}

\begin{table}[t]
  \centering
  \caption{Per-class statistics of \dataset, showing the inter-procedural (FCG) and intra-procedural (CFG) graph structures for benign applications, the aggregate malicious set, and its constituent malware families. All node and edge counts are per-graph averages.}
  \label{tab:dataset_stats}
  \small
  \setlength{\tabcolsep}{8pt}
  \begin{tabular}{@{}lrrrrr@{}}
    \toprule
    \textbf{Class}                    & \textbf{\# Apps} & \multicolumn{2}{c}{\textbf{FCG (per app)}} & \multicolumn{2}{c}{\textbf{CFG (per function)}} \\
    \cmidrule(lr){3-4} \cmidrule(lr){5-6}
                                      &                  & \textbf{Avg.\ Nodes} & \textbf{Avg.\ Edges} & \textbf{Avg.\ Nodes} & \textbf{Avg.\ Edges} \\
    \midrule
    Benign                            & 449,320          &  791.54             & 1,414.51             & 12.17                & 13.94                \\
    \midrule
    Malicious (total)                 &  50,661          &  266.48             &   491.67             & 12.29                & 14.94                \\
    \quad Adware                      &  25,633          &  337.38             &   615.80             & 12.25                & 14.94                \\
    \quad Grayware                    &  27,018          &  141.76             &   259.13             & 12.73                & 15.62                \\
    \quad Tool                        &     377          &  300.79             &   531.73             & 11.49                & 13.83                \\
    \quad Downloader                  &     159          & 1,453.33            & 3,370.85             & 12.14                & 14.56                \\
    \quad Spyware                     &      46          &  439.07             &   793.15             & 11.27                & 13.71                \\
    \quad Backdoor                    &      45           &  238.44             &   411.71             & 12.35                & 14.92                \\
    \quad Clicker                     &      94          &  112.04             &   157.67             & 12.18                & 14.26                \\
    \bottomrule
  \end{tabular}
\end{table}

\begin{table*}[htbp]
  \centering
  \caption{Detailed Function Call Graph (FCG) Statistics by Malware Class.}
  \label{tab:fcg_stats}
  {\footnotesize
    \setlength{\tabcolsep}{4pt}
    \begin{tabular}{@{}lrrrrrrrrrrrrr@{}}
      \toprule
      \multirow{2}{*}{\textbf{Class}} & \multirow{2}{*}{\textbf{\# FCGs}} & \multicolumn{3}{c}{\textbf{Nodes}} & \multicolumn{3}{c}{\textbf{Edges}} & \multicolumn{3}{c}{\textbf{Average Degree}} & \multicolumn{3}{c}{\textbf{Density}}                                                               \\
      \cmidrule(lr){3-5} \cmidrule(lr){6-8} \cmidrule(lr){9-11} \cmidrule(lr){12-14}
                                      &                                   & Min                                & Mean                               & Max                                         & Min                                  & Mean     & Max    & Min  & Mean & Max  & Min  & Mean & Max  \\
      \midrule
      Grayware                        & 27,018                            & 2                                  & 141.76                             & 9,901                                       & 4                                    & 259.13   & 24,354 & 0    & 2.02 & 5.60 & 0.01 & 0.10 & 1.00 \\
      Adware                          & 25,633                            & 2                                  & 337.38                             & 7,241                                       & 4                                    & 615.80   & 17,700 & 0    & 3.04 & 8.62 & 0.02 & 0.03 & 1.00 \\
      Tool                            & 377                               & 9                                  & 300.79                             & 2,182                                       & 7                                    & 531.73   & 4,470  & 1.54 & 2.91 & 5.69 & 0.01 & 0.02 & 0.19 \\
      Downloader                      & 159                               & 3                                  & 1,453.33                           & 12,326                                      & 2                                    & 3,370.85 & 31,292 & 1.17 & 3.01 & 5.26 & 0.02 & 0.04 & 0.67 \\
      Clicker                         & 94                                & 3                                  & 112.04                             & 706                                         & 2                                    & 157.67   & 1,221  & 1.33 & 2.25 & 4.00 & 0.01 & 0.09 & 0.67 \\
      Spyware                         & 46                                & 32                                 & 439.07                             & 2,320                                       & 35                                   & 793.15   & 4,837  & 2.00 & 2.85 & 4.67 & 0.01 & 0.03 & 0.07 \\
      Backdoor                        & 45                                & 19                                 & 238.44                             & 1,626                                       & 18                                   & 411.71   & 3,302  & 1.84 & 2.65 & 4.22 & 0.01 & 0.03 & 0.11 \\
      \bottomrule
    \end{tabular}
  }
\end{table*}

\begin{table*}[htbp]
  \centering
  \caption{Detailed Control Flow Graph (CFG) Statistics by Malware Class.}
  \label{tab:cfg_stats}
  {\footnotesize
    \setlength{\tabcolsep}{4pt}
    \begin{tabular}{@{}lrrrrrrrrrrrrr@{}}
      \toprule
      \multirow{2}{*}{\textbf{Class}} & \multirow{2}{*}{\textbf{\# CFGs}} & \multicolumn{3}{c}{\textbf{Nodes}} & \multicolumn{3}{c}{\textbf{Edges}} & \multicolumn{3}{c}{\textbf{Average Degree}} & \multicolumn{3}{c}{\textbf{Density}}                                                            \\
      \cmidrule(lr){3-5} \cmidrule(lr){6-8} \cmidrule(lr){9-11} \cmidrule(lr){12-14}
                                      &                                   & Min                                & Mean                               & Max                                         & Min                                  & Mean  & Max   & Min  & Mean & Max   & Min  & Mean & Max  \\
      \midrule
      Adware                          & 3,838,620                         & 2                                  & 12.25                              & 1,281                                       & 1                                    & 14.94 & 2,238 & 0.20 & 2.20 & 8.50  & 0.02 & 0.39 & 2.83 \\
      Grayware                        & 1,750,737                         & 2                                  & 12.73                              & 932                                         & 1                                    & 15.62 & 1,455 & 0.20 & 2.20 & 26.00 & 0.02 & 0.38 & 8.50 \\
      Downloader                      & 131,311                           & 2                                  & 12.14                              & 2,093                                       & 1                                    & 14.56 & 2,117 & 0.33 & 2.18 & 4.80  & 0.01 & 0.39 & 2.33 \\
      Tool                            & 50,863                            & 2                                  & 11.49                              & 591                                         & 1                                    & 13.83 & 742   & 0.29 & 2.16 & 5.77  & 0.01 & 0.39 & 2.00 \\
      Spyware                         & 9,827                             & 2                                  & 11.27                              & 718                                         & 1                                    & 13.71 & 917   & 0.40 & 2.21 & 3.69  & 0.01 & 0.38 & 1.00 \\
      Backdoor                        & 4,718                             & 3                                  & 12.35                              & 369                                         & 1                                    & 14.92 & 582   & 0.50 & 2.15 & 3.38  & 0.01 & 0.38 & 1.00 \\
      Clicker                         & 3,741                             & 2                                  & 12.18                              & 199                                         & 1                                    & 14.26 & 249   & 0.50 & 2.10 & 4.51  & 0.01 & 0.38 & 1.00 \\
      \bottomrule
    \end{tabular}
  }
\end{table*}

\section{Extended Temporal Analysis}
\label{sec:extended_temporal}

\begin{figure}[htbp]
  \centering
  \begin{minipage}[b]{0.62\textwidth}
    \centering
    \includegraphics[width=\linewidth]{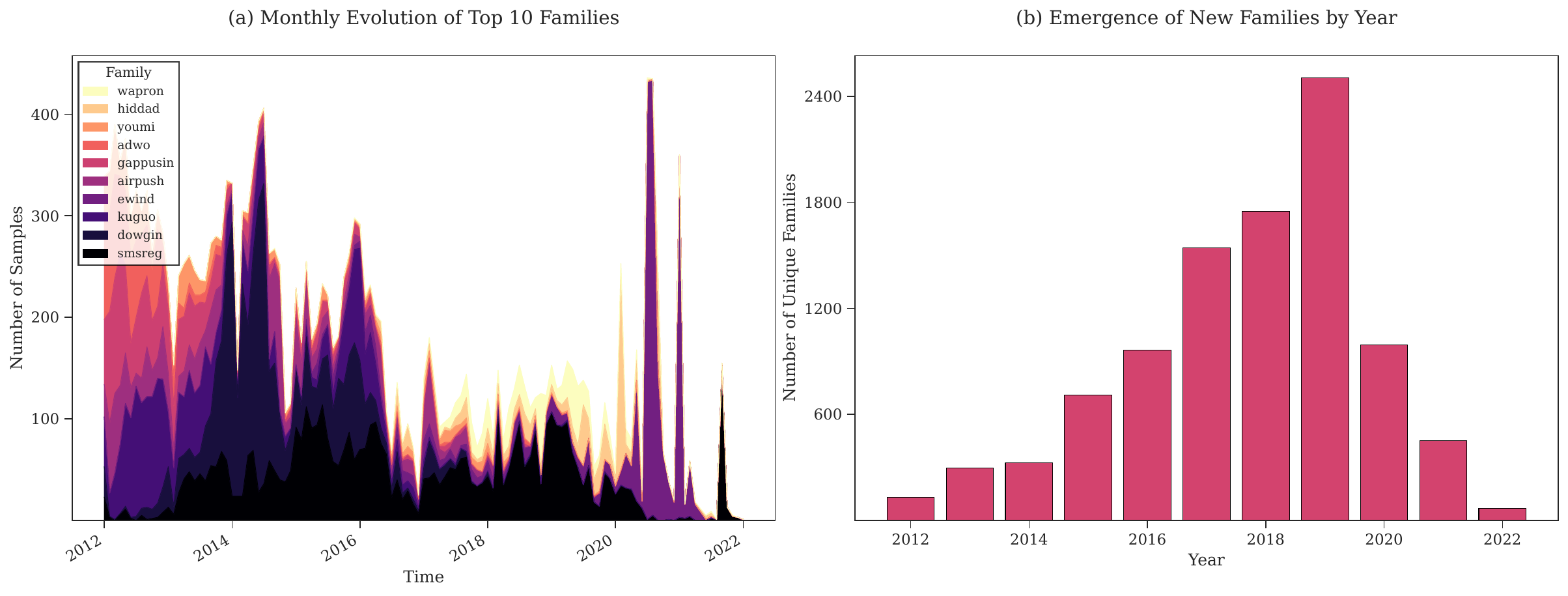}
    \subcaption{Malware family evolution. Left: monthly distribution of top-10 families (\textit{dowgin}, \textit{smsreg}, etc.); Right: annual count of unique families.}
    \label{fig:temporal_evolution}
  \end{minipage}\hfill
  \begin{minipage}[b]{0.36\textwidth}
    \centering
    \includegraphics[width=\linewidth]{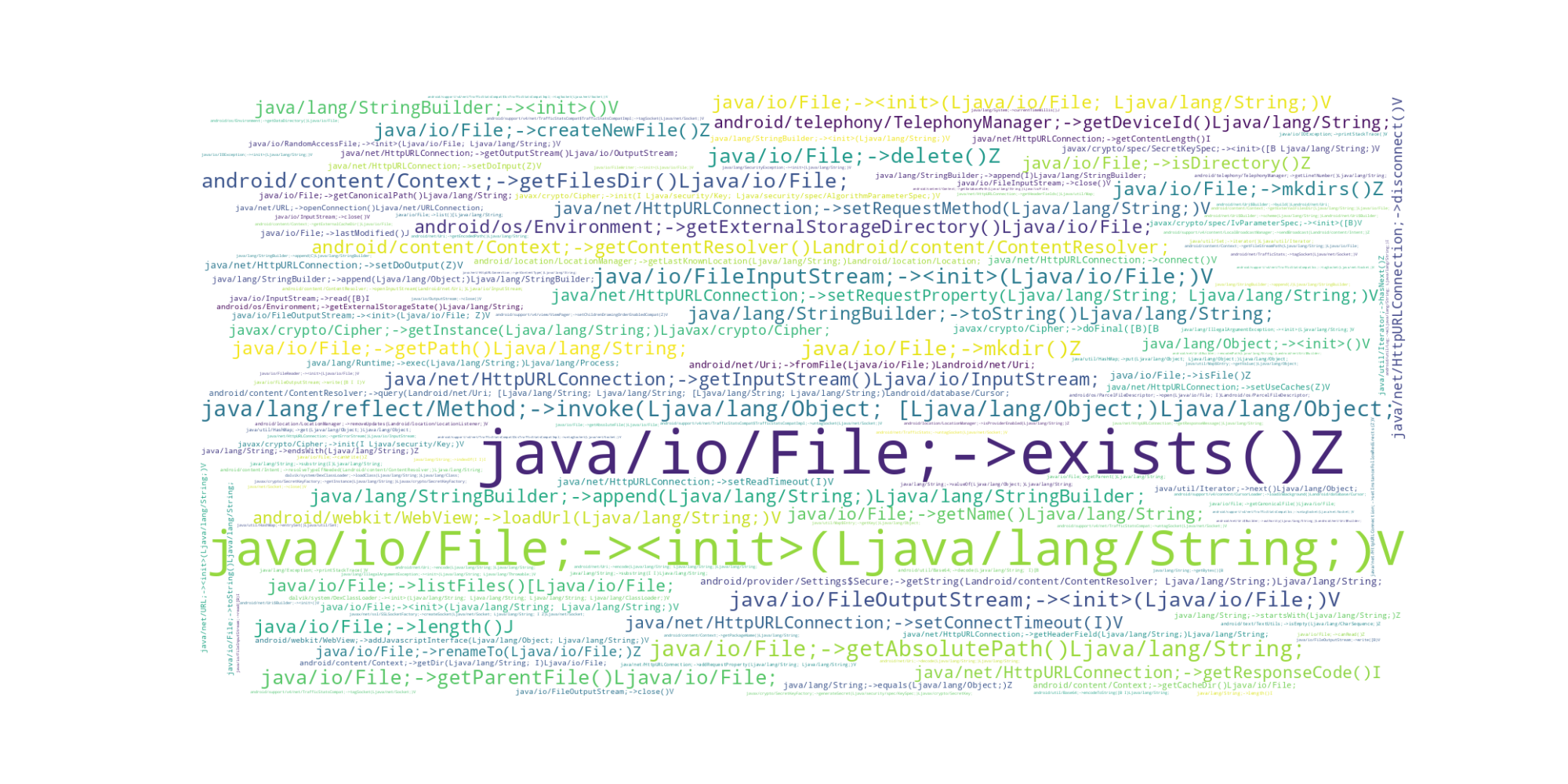}
    \subcaption{API usage word cloud; font size reflects invocation frequency across all 500K \dataset applications.}
    \label{fig:api_wordcloud}
  \end{minipage}
  \caption{Dataset-level temporal and functional fingerprints in \dataset. \textbf{(a)} Top-10 malware families fluctuate over 2012 to 2022 (left panel of (a)) while the count of new unique families grows exponentially before peaking around 2019 to 2020 (right panel of (a)), reflecting the security arms race. \textbf{(b)} Frequent API function names highlight both common utility APIs (\textit{java/io/File}, \textit{HttpURLConnection}) and security-sensitive platform APIs (\textit{TelephonyManager}), the latter being primary targets of malicious exploitation.}
  \label{fig:family_and_api}
\end{figure}

This section reports temporal trends in both Function Call Graph (FCG) and Control Flow Graph (CFG) structural properties over 2012--2022, supplementing the structural comparison in Section~\ref{sec:structural_comparison} of the main paper. We also include the dataset-level family-evolution and API-usage fingerprints (Figure~\ref{fig:family_and_api}).

\paragraph{Implications for detection.} The temporal patterns observed in the main text have direct implications for detector design. The observed structural divergence between malware and benign apps suggests that density-based metrics serve as increasingly powerful discriminative features over time: malware's tendency to maintain high structural density while benign software favors modularity provides a fundamental architectural signature that persists across temporal evolution. This finding underscores the value of \dataset's longitudinal design for understanding persistent malware characteristics amid evolving threat landscapes, and motivates feature-engineering choices that emphasize density and hierarchy over flat-graph aggregates.

\begin{figure*}[htbp]
  \centering
  \begin{subfigure}[b]{0.32\textwidth}
    \centering
    \includegraphics[width=\textwidth]{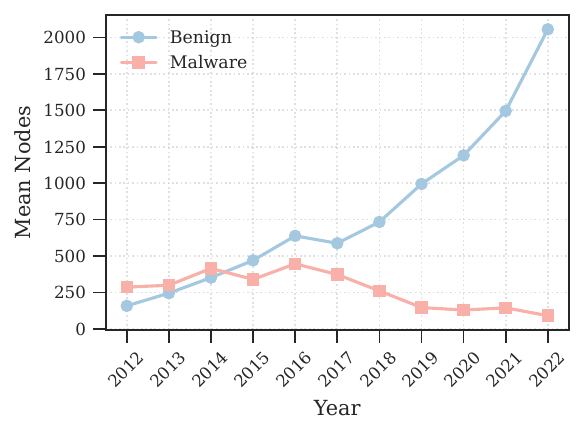}
    \caption{Mean Nodes Count}
    \label{fig:fcg_mean_nodes_trend}
  \end{subfigure}
  \hfill
  \begin{subfigure}[b]{0.32\textwidth}
    \centering
    \includegraphics[width=\textwidth]{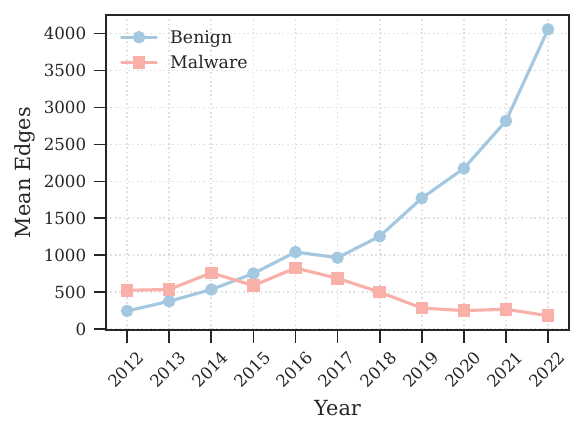}
    \caption{Mean Edges Count}
    \label{fig:fcg_mean_edges_trend}
  \end{subfigure}
  \hfill
  \begin{subfigure}[b]{0.32\textwidth}
    \centering
    \includegraphics[width=\textwidth]{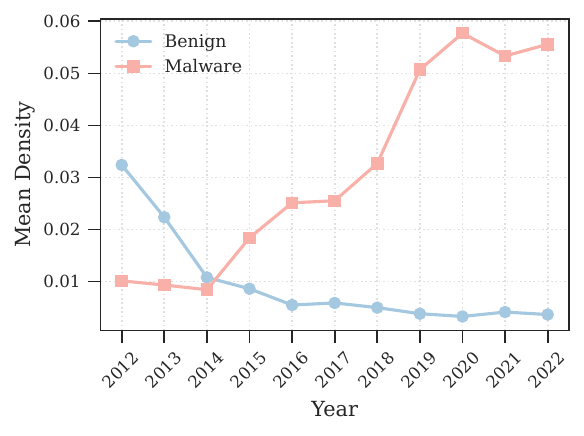}
    \caption{Mean Density Count}
    \label{fig:fcg_mean_density_trend}
  \end{subfigure}
  \caption{Trends in Function Call Graph (FCG) structural properties over time, showing mean nodes, mean edges, and mean density.}
  \label{fig:fcg_trends_main}
\end{figure*}

\begin{figure*}[htbp]
  \centering
  \begin{subfigure}[b]{0.32\textwidth}
    \centering
    \includegraphics[width=\textwidth]{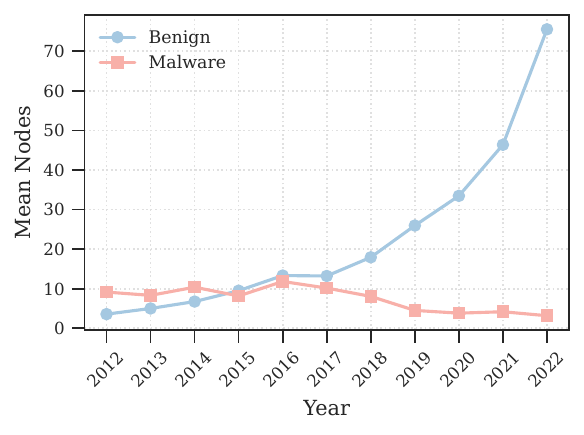}
    \caption{Mean Nodes Count}
    \label{fig:cfg_mean_nodes_trend}
  \end{subfigure}
  \hfill
  \begin{subfigure}[b]{0.32\textwidth}
    \centering
    \includegraphics[width=\textwidth]{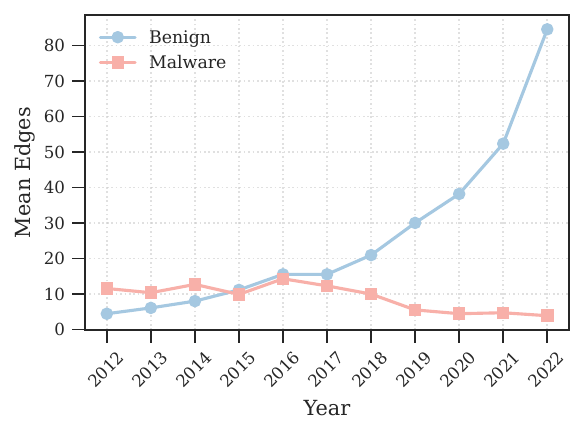}
    \caption{Mean Edges Count}
    \label{fig:cfg_mean_edges_trend}
  \end{subfigure}
  \hfill
  \begin{subfigure}[b]{0.32\textwidth}
    \centering
    \includegraphics[width=\textwidth]{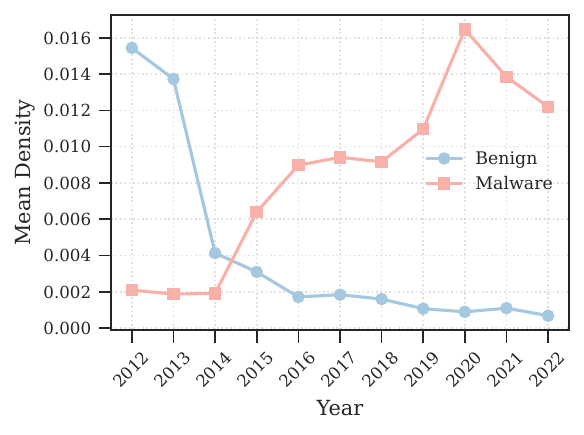}
    \caption{Mean Density Count}
    \label{fig:cfg_mean_density_trend}
  \end{subfigure}
  \caption{Trends in Control Flow Graph (CFG) structural properties over time, showing mean nodes, mean edges, and mean density.}
  \label{fig:cfg_trends}
\end{figure*}

Examining the temporal evolution of CFGs (Figure~\ref{fig:cfg_trends}), we observe that benign applications show accelerating growth in both nodes and edges, particularly after 2016. This indicates increasing intra-procedural complexity. In contrast, malware CFGs remain small and relatively stable over time.

Analyzing graph density, benign CFGs become sparser over time despite their growth in size, suggesting a trend towards more structured and less complex functions. Conversely, malware CFG density remains consistently higher than that of benign software after 2014. This points to more intricate internal logic within smaller functions, possibly as a result of obfuscation techniques.

\section{Visual Analysis of Concept Drift}
\label{sec:visual_drift}

To complement the model-side concept-drift evidence in the main paper, we directly probe \dataset's feature distributions for drift. Following standard distributional drift analysis~\cite{pendlebury2019tesseract}, we (i) project per-app aggregate features to two dimensions via t-SNE and (ii) compute per-feature Jeffreys divergence relative to the earliest year. Both analyses are computed on a stratified subsample of $\sim$30K apps (1{,}500 malware + 1{,}500 benign per year) over 2012--2022.

\begin{figure}[t]
  \centering
  \includegraphics[width=\linewidth]{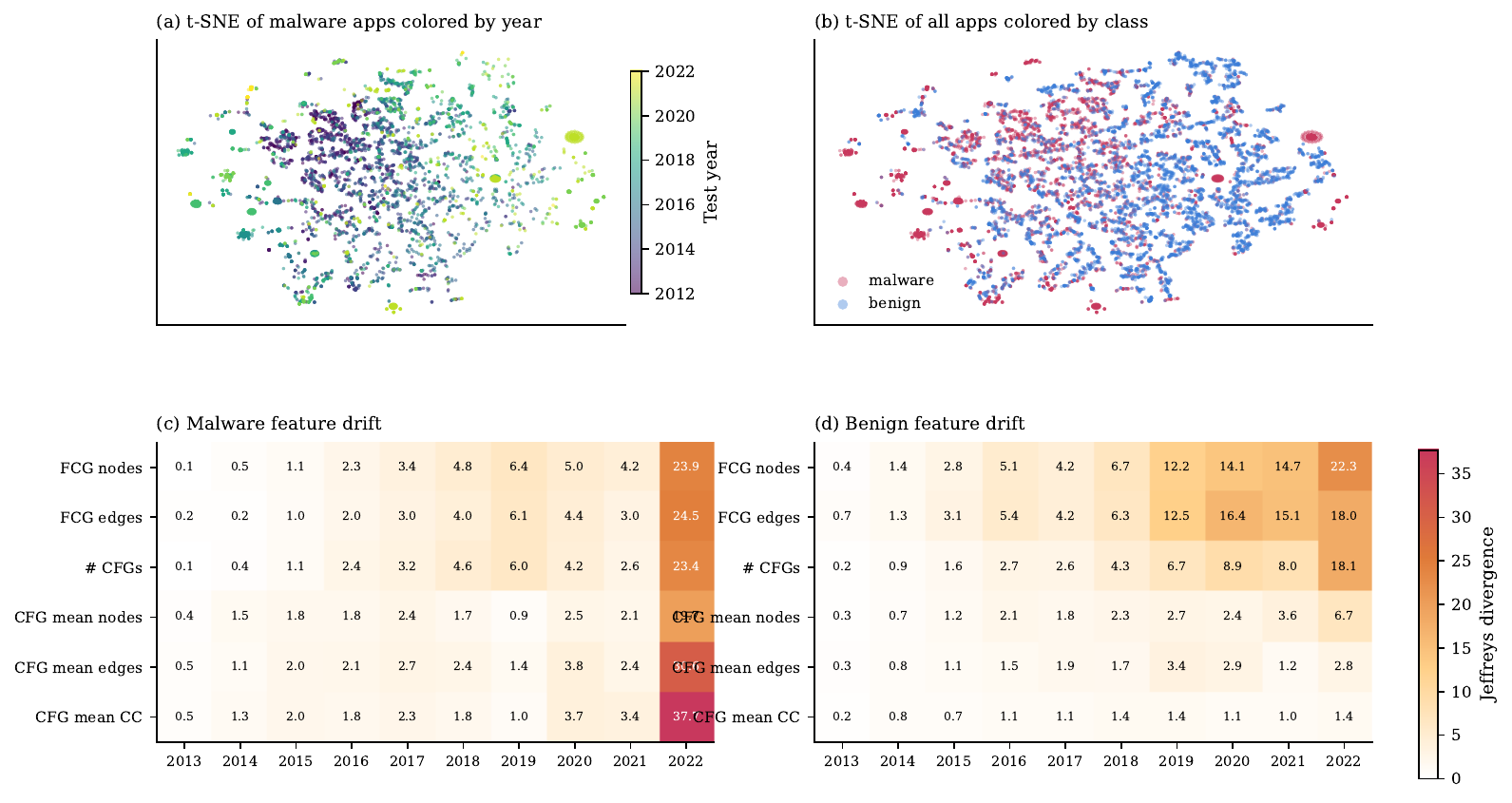}
  \caption{Visual analysis of concept drift in \dataset. Top: t-SNE on six aggregate features colored by year (a) and class (b). Bottom: per-feature Jeffreys divergence relative to 2012 for malware (c) and benign (d) apps.}
  \label{fig:drift_visual}
\end{figure}

Figure~\ref{fig:drift_visual} consolidates the two analyses into one panel. The top row shows the t-SNE projection: in panel~(a), malware apps colored by release year reveal distinct year-cohorts clustering separately, with later-year apps occupying regions early-year apps do not, providing direct visual evidence of structural drift. Panel~(b) overlays malware (red) and benign (blue); the two classes share large portions of the manifold, yet malware concentrates in compact clusters that benign apps do not occupy, showing that even simple aggregate features carry discriminative signal. The bottom row reports Jeffreys divergence $J(P_{\text{year}} \,\|\, P_{2012})$ for each of the six aggregate features. All features drift monotonically away from 2012, confirming that \dataset captures real, gradual distribution shift. Benign FCG-level features (panel~d, top three rows) drift faster ($J{>}10$ from 2019 onward), reflecting the rapid growth of benign app size observed in Section~\ref{sec:structural_comparison} of the main paper, while malware retains relatively stable CFG-level distributions until the small-sample 2022 endpoint (only 67 confirmed malware at VT$\geq$15) where divergence values should be interpreted alongside sample budget rather than as a stronger drift event.

\section{Complete Experimental Configuration}
\label{sec:experimental_setup}

This section provides comprehensive experimental details that supplement the experimental setup described in Section~\ref{sec:application} of the main paper. All technical specifications and hyperparameter configurations are documented here to ensure full reproducibility.

\subsection{Detailed Model Architectures}
The model architectures detailed in the main paper are expanded here with specific implementation details. For the baseline GNN models (GCN, GAT, GIN, GraphSAGE) used for graph-level classification, we employ a consistent architecture comprising two GNN layers with ReLU activation functions, followed by global mean pooling and a fully connected classification layer with logarithmic softmax output. Dropout regularization is applied uniformly across baseline models.

Hi-GNN employs a dual-encoder architecture with separate GNN encoders for CFGs and FCGs. Each encoder uses the same two-layer structure as the baselines, with learned representations integrated through concatenation followed by a linear transformation before the final classification layer.

\subsection{Baseline Methods}
We compare the performance of Hi-GNN against several widely-adopted GNN baseline models. These models are standard benchmarks for graph classification tasks:
\begin{itemize}
  \item \textbf{Graph Convolutional Network (GCN)} \cite{kipf2017semi}: Aggregates information from immediate neighbors using spectral graph convolutions.
  \item \textbf{Graph Attention Network (GAT)} \cite{velivckovic2018graph}: Employs attention mechanisms to assign different weights to neighbor contributions during feature aggregation.
  \item \textbf{Graph Isomorphism Network (GIN)} \cite{xu2018powerful}: A GNN variant designed to be as expressive as the Weisfeiler-Lehman graph isomorphism test, utilizing MLPs for node feature transformation and sum aggregation.
  \item \textbf{GraphSAGE} \cite{hamilton2017inductive}: An inductive learning framework that samples a fixed number of neighbors and applies various aggregator functions (e.g., mean, LSTM, pooling) to combine neighbor features.
  \item \textbf{GraphGPS} \cite{rampavsek2022recipe}: A modern graph transformer that combines local GIN message passing with global multi-head self-attention. We use the recipe variant with a 128-dim hidden representation and Performer-based linear attention; FCG nodes are encoded by aggregating their per-function CFG features. Trained with the same Adam configuration as the other baselines.
  \item \textbf{Hi-GNN}: A hierarchical graph neural network designed to process both local Control Flow Graphs (CFGs) and the global Function Call Graph (FCG). We use GCN as the GNN encoder for both CFG and FCG.
\end{itemize}
All baseline models were implemented using the PyTorch Geometric library \cite{Fey/Lenssen/2019}.

\subsection{Hyperparameters}
\begin{table}[!t]
  \centering
  \caption{Hyperparameter settings for GNN models. Hidden dimension is fixed at 128 across all models (baselines, GraphGPS, and Hi-GNN encoders) for matched-capacity comparison; remaining values are searched on validation performance.}
  \label{tab:hyperparam_settings}
  \footnotesize
  \setlength{\tabcolsep}{4pt}
  \begin{tabular}{@{}p{2cm}p{2.5cm}p{2.5cm}@{}}
    \toprule
    \textbf{Model}                  & \textbf{Parameter} & \textbf{Value Range} \\
    \midrule
    \multirow{6}{*}{Baseline GNNs}  & GNN Layers         & 1-3                  \\
                                    & Hidden Dimension   & 128 (fixed)          \\
                                    & Batch Size         & 32, 64, 128          \\
                                    & Pooling Layer      & Global Mean          \\
                                    & Learning Rate      & 0.001, 0.01          \\
                                    & Number of Epochs   & 50, 100, 200         \\
                                    & Dropout Rate       & 0.4, 0.5, 0.6        \\
    \midrule
    \multirow{6}{*}{Hi-GNN Encoder} & GNN Layers         & 1-3                  \\
                                    & Hidden Dimension   & 128 (fixed)          \\
                                    & Batch Size         & 32, 64, 128, 256     \\
                                    & Pooling Layer      & Global Mean          \\
                                    & Learning Rate      & 0.001, 0.01          \\
                                    & Number of Epochs   & 50, 100, 200         \\
                                    & Dropout Rate       & 0.4, 0.5, 0.6        \\
    \bottomrule
  \end{tabular}
\end{table}
We test on the hyperparameter settings shown in Table~\ref{tab:hyperparam_settings} and select the best configuration based on validation performance. All models were trained using the Adam optimizer with cross-entropy loss for up to 200 epochs, with early stopping based on validation Macro F1. We use PR-AUC, Macro F1, Precision, and Recall as evaluation metrics, focusing on Macro F1 due to the inherent class imbalance ($\sim$10\% malware samples). Each experiment is repeated three times with different random seeds, and we report the mean and standard deviation to ensure statistical reliability.

\subsection{Hi-GNN Architecture Details}
The Hi-GNN architecture compressed in Section~\ref{sec:application} is detailed here. Hi-GNN employs a dual-encoder design with two distinct two-layer GNN encoders---one for CFG nodes (intra-procedural) and one for FCG nodes (inter-procedural)---each followed by ReLU activations and global mean pooling. The CFG encoder produces per-function embeddings; for each function node in the FCG, this CFG embedding is concatenated with structural features (in/out-degree) and then passed through the FCG encoder. The fused multi-level representations are then passed through a linear transformation before the final classification layer with logarithmic softmax output, enabling Hi-GNN to capture both intra-procedural control flow patterns and inter-procedural functional dependencies simultaneously.

\subsection{Non-GNN Baseline: XGBoost on Graph Statistics}
\label{sec:xgboost_baseline}
To probe whether hierarchical GNN modeling is necessary or whether simple aggregate statistics already suffice, we train an XGBoost classifier on the 6 per-app aggregate features (\texttt{n\_functions}, \texttt{n\_calls}, \texttt{n\_cfgs}, \texttt{mean\_cfg\_nodes}, \texttt{mean\_cfg\_edges}, \texttt{mean\_cfg\_cc}). To keep the experiment lightweight we evaluate on a stratified balanced subsample (1,500 malware $+$ 1,500 benign per year, 30K apps over 2012--2022); absolute Macro F1 values are therefore not directly comparable to Table~\ref{tab:benchmark_combined} (which uses the imbalanced $\sim$10\% full cohort) but they isolate the contribution of structural features alone.

\textbf{Static IID (2012, balanced).} XGBoost reaches Macro F1 $=0.829\pm0.014$ and PR-AUC $=0.912\pm0.014$ across 3 random seeds with a 70/15/15 split; logistic regression on standardized features reaches $0.756\pm0.010$ Macro F1.

\textbf{Temporal (Train 2012 $\rightarrow$ Test 2013--2016, yearly).} XGBoost Macro F1 by year: 2013$=0.694$, 2014$=0.628$, 2015$=0.504$, 2016$=0.516$; yearly-averaged AUT(F1) for 2012$\rightarrow$2016 is approximately $0.585$, comparable to GraphGPS ($0.582$, Table~\ref{tab:benchmark_combined}) but well below Hi-GNN ($0.715$).

\textbf{Interpretation.} Simple statistical features carry substantial discriminative signal under balanced static evaluation, consistent with the medium Cohen's $d{=}0.48$ for CFG cyclomatic complexity (Table~\ref{tab:stat_significance}). However, XGBoost degrades sharply under temporal drift, while Hi-GNN's hierarchical aggregation sustains stronger long-horizon performance ($0.715$ vs.\ $0.585$ at the 4-year horizon, a $0.13$ AUT(F1) gap). The advantage of hierarchical modeling therefore manifests most strongly in temporal robustness, on top of the static-IID gains over flat GNN baselines reported in Table~\ref{tab:benchmark_combined}. A full ablation (FCG-only, randomized hierarchy, parameter-matched controls) is left to follow-up work.

\section{Sensitive API Filter}
\label{sec:sensitive_api}

To reduce noise in the FCG and focus on security-relevant interactions, we retain edges whose endpoints match one of 24 sensitive API patterns spanning 10 categories (Table~\ref{tab:sensitive_api_filter}). The patterns are matched against the full dotted class/method names produced by Androguard. The complete pattern list is also released in \texttt{apk\_to\_higraph.py} of the public preprocessing pipeline.

\begin{table}[t]
  \centering
  \caption{Sensitive API filter applied during FCG construction. Edges in the FCG are kept when either endpoint matches one of these patterns; remaining utility-only edges are pruned.}
  \label{tab:sensitive_api_filter}
  \footnotesize
  \setlength{\tabcolsep}{4pt}
  \begin{tabular}{@{}lcl@{}}
    \toprule
    \textbf{Category} & \textbf{\#} & \textbf{Example patterns} \\
    \midrule
    SMS              & 3 & \texttt{android/telephony/SmsManager}, \texttt{sendTextMessage}, \texttt{sendMultipartTextMessage} \\
    Networking       & 3 & \texttt{java/net/HttpURLConnection}, \texttt{org/apache/http}, \texttt{java/net/Socket} \\
    Dynamic loading  & 2 & \texttt{dalvik/system/DexClassLoader}, \texttt{java/lang/ClassLoader} \\
    Cryptography     & 3 & \texttt{javax/crypto/Cipher}, \texttt{SecretKeySpec}, \texttt{MessageDigest} \\
    File I/O         & 2 & \texttt{java/io/File}, \texttt{java/io/FileOutputStream} \\
    Location         & 2 & \texttt{android/location/LocationManager}, \texttt{getLastKnownLocation} \\
    Device admin     & 2 & \texttt{android/app/admin/DevicePolicyManager}, \texttt{lockNow} \\
    Reflection/exec  & 3 & \texttt{java/lang/reflect/Method}, \texttt{Runtime/exec}, \texttt{ProcessBuilder} \\
    Telephony        & 2 & \texttt{android/telephony/TelephonyManager}, \texttt{getDeviceId} \\
    WebView          & 2 & \texttt{android/webkit/WebView}, \texttt{loadUrl} \\
    \midrule
    \textbf{Total}   & \textbf{24} & across 10 categories \\
    \bottomrule
  \end{tabular}
\end{table}

\section{Known Limitations}
\label{sec:limitations}

We document four known limitations to guide downstream use of \dataset.

\paragraph{Sample budget for 2021--2022.} The yearly counts decline sharply after 2020 (Table~\ref{tab:yearly_stats}: $\sim$22K apps in 2021, only 825 in 2022). This is \emph{not} a collection gap but a consequence of our strict labeling threshold: an application is labeled malicious only when at least 15 VirusTotal engines flag it~\cite{zhang2020enhancing}. As antivirus consensus accumulates over time~\cite{allix2016androzoo}, recent samples gradually cross the threshold; for 2022, only 67 apps currently meet it (and 7 for 2023, excluded from the release). Drift metrics for these years should therefore be interpreted alongside their sample budget rather than as standalone trend evidence. We commit to periodic re-labeling and version-controlled releases as VirusTotal coverage matures for recent years.

\paragraph{Single-platform coverage.} \dataset focuses on Android because AndroZoo provides the largest curated repository with reliable temporal metadata at this scale. The hierarchical CFG/FCG representation, however, is platform-agnostic: the same construction pipeline can in principle be applied to Windows PE or Linux ELF binaries by substituting Androguard with a comparable disassembler (e.g., Ghidra, IDA, or radare2). We leave cross-platform extension to future work.

\paragraph{Label noise and threshold choice.} The VT$\geq$15 threshold is a common but conservative choice~\cite{zhang2020enhancing}; users requiring higher recall on borderline samples can re-derive labels from the released VirusTotal reports at thresholds of their choosing. AVClass2~\cite{sebastian2020avclass2} family labels inherit any noise from upstream antivirus naming conventions.

\paragraph{Static analysis only.} Our representation captures static program structure and does not model dynamic behavior (e.g., runtime API invocations, network traffic, or sandbox traces). Detectors that rely on dynamic features will need to be combined with complementary dynamic datasets such as CICMalDroid~\cite{mahdavifar2020dynamic}.

\section{Computational Resources}
Our data preprocessing pipeline, including downloading, preprocessing, decompilation, and label extraction, was performed on the Neptune cluster running RHEL 8.8. Each node in this cluster features dual AMD EPYC 9354 processors at 3.25GHz with 32 cores, 768GB of 4800MHz DDR5-RAM in twelve-channel configuration. We also have 20TB of project storage for storing the dataset.

The model training was conducted on the Saturn cluster, also running RHEL 8.8. The compute nodes are equipped with AMD EPYC 9254 processors running at 2.9GHz with 24 cores, 192GB of 4800MHz DDR5-RAM in twelve-channel configuration, dual 1.92TB NVMe SSD drives, and two NVIDIA L40 GPUs per node, each featuring 48GB of memory.

\section{Dataset Documentation}
\label{sec:dataset_documentation}

\subsection{Hosted URLs}

\paragraph{Project page.} \url{https://higraph.org} --- interactive explorer for FCG/CFG samples and per-family statistics.

\paragraph{Hugging Face dataset.} \url{https://huggingface.co/datasets/hzcheney/Hi-Graph} --- parquet-format release with year-stratified shards.

\paragraph{GitHub code.} \url{https://github.com/hzcheney/HiGraph} --- preprocessing pipeline (\texttt{apk\_to\_higraph.py}), Hi-GNN / GraphGPS / baseline implementations, and evaluation harness.

\paragraph{Croissant metadata.} \url{https://huggingface.co/api/datasets/hzcheney/Hi-Graph/croissant} --- auto-generated by Hugging Face, covering core fields, distribution, recordSet, and Responsible AI fields (data limitations, biases, use cases).

\subsection{Accessibility and Reproducibility}

The dataset is publicly available on Hugging Face under \texttt{hzcheney/Hi-Graph}, with a version-controlled release plan tied to GitHub tags. The interactive explorer at \url{https://higraph.org} provides curated CFG/FCG samples and per-family statistics for fast browsing without downloading the full 6\,GB corpus. We commit to long-term maintenance: VirusTotal labels are periodically re-queried as antivirus consensus matures for recent years, and updated releases are versioned to ensure reproducibility of all reported numbers.

\subsection{License}

The dataset and code are released under the Creative Commons Attribution-NonCommercial-ShareAlike 4.0 International License (CC-BY-NC-SA 4.0, SPDX identifier \texttt{CC-BY-NC-SA-4.0}). This license allows remixing, adapting, and building upon our work for non-commercial purposes with appropriate credit and ShareAlike terms.

\subsection{Author Statement}

The authors bear full responsibility for any rights violations or licensing concerns related to the released artifacts. Any feedback or issue reports can be filed via the GitHub repository's issue tracker; we will respond and update the release accordingly.

\end{document}